\documentclass[journal,twoside]{IEEEtran}

\usepackage{cite}
\ifCLASSINFOpdf
\usepackage[pdftex]{graphicx}
\else
\usepackage[dvips]{graphicx}
\fi
\ifCLASSOPTIONcompsoc
\usepackage[caption=false,font=normalsize,labelfont=sf,textfont=sf]{subfig}
\else
\usepackage[caption=false,font=footnotesize]{subfig}
\fi
\graphicspath{{./figures/}{}}
\DeclareGraphicsExtensions{.pdf, .png}
\usepackage{array}
\usepackage{multirow}
\usepackage{makecell}
\usepackage{booktabs}
\usepackage[cmex10]{amsmath}
\usepackage[]{amssymb}
\usepackage{siunitx}
\usepackage{algorithm}
\usepackage{algorithmic}

\usepackage{hyperref}
\usepackage{url}

\usepackage{color}
\usepackage{colortbl}
\usepackage{xcolor}

\newcommand{\deepen}{\cellcolor{lightgray}\textbf}
\newcommand{\idmat}{\mathbf{I}}

\newcommand{\real}{\mathbb{R}}
\newcommand{\graph}{\mathcal{G}}
\newcommand{\vertset}{\mathcal{V}}
\newcommand{\vertelm}{v}
\newcommand{\nvert}{n}
\newcommand{\ivert}{i}
\newcommand{\jvert}{j}
\newcommand{\dvert}{d}
\newcommand{\edgeset}{\mathcal{E}}
\newcommand{\edgeelm}{e}
\newcommand{\nedge}{m}
\newcommand{\ftrmat}{\mathbf{X}}
\newcommand{\ftrvec}{\mathbf{x}}
\newcommand{\adjmat}{\mathbf{A}}
\newcommand{\degmat}{\mathbf{D}}
\newcommand{\laplamat}{\mathbf{L}}
\newcommand{\gconv}{*_\graph}
\newcommand{\conv}{*}
\newcommand{\wgtmat}{\mathbf{W}}
\newcommand{\biasvec}{\mathbf{b}}
\newcommand{\actfun}{f}
\newcommand{\dataset}{\mathcal{D}}
\newcommand{\ndataset}{{n^\text{ds}}}
\newcommand{\idataset}{k}
\newcommand{\nband}{{n^\text{band}}}
\newcommand{\nsband}{n^\text{sband}}
\newcommand{\iband}{i}
\newcommand{\jband}{j}
\newcommand{\hidmat}{\hat{\mathbf{H}}}
\newcommand{\scrvec}{\hat{\mathbf{s}}}
\newcommand{\bsgtvec}{\mathbf{s}}
\newcommand{\mskvec}{\hat{\mathbf{M}}}
\newcommand{\adjthres}{\theta^\text{adj}}
\newcommand{\scrthres}{\theta^\text{band}}
\newcommand{\nbase}{{n^\text{base}}}
\newcommand{\istage}{i}
\newcommand{\ipatch}{i}
\newcommand{\batchsize}{n^\text{batch}}
\newcommand{\height}{h}
\newcommand{\width}{w}
\newcommand{\nchannel}{C}
\newcommand{\nclass}{n^\text{class}}
\newcommand{\half}{\frac{1}{2}}
\newcommand{\dband}{\height\width}
\newcommand{\dhid}{256}
\newcommand{\bin}{\phi}
\newcommand{\operator}{\mathcal{F}}
\newcommand{\patch}{x}
\newcommand{\prbvec}{\hat{\mathbf{y}}}
\newcommand{\clsgtvec}{y}
\newcommand{\loss}{\mathcal{L}}
\newcommand{\iepoch}{t}
\newcommand{\nepoch}{T}
\newcommand{\weight}{\lambda}
\newcommand{\lr}{\alpha}
\newcommand{\mlr}{\beta}

\DeclareMathOperator{\relu}{ReLU}
\DeclareMathOperator{\sigmoid}{\sigma}
\DeclareMathOperator{\softmax}{Softmax}
\DeclareMathOperator{\bn}{BN}

\renewcommand{\thesubsection}{\Alph{subsection}}
\renewcommand\thetable{\Roman{table}}

\hypersetup{
    colorlinks=true,
    linkcolor=red,
    citecolor=red,
    urlcolor=magenta,
}

\begin{document}

\title{Multi-Teacher Multi-Objective Meta-Learning\\for Zero-Shot Hyperspectral Band Selection}


\author{Jie~Feng,~\IEEEmembership{Senior~Member,~IEEE,}
        Xiaojian~Zhong,
        Di~Li,
        Weisheng~Dong,~\IEEEmembership{Member,~IEEE,}
        Ronghua~Shang,~\IEEEmembership{Senior~Member,~IEEE,}
        and~Licheng~Jiao,~\IEEEmembership{Fellow,~IEEE}%
\thanks{J. Feng, X. Zhong, D. Li, W. Dong, R. Shang and L. Jiao are with the Key Laboratory of Intelligent Perception and Image Understanding of Ministry of Education of China, Xidian University, Xi'an 710071, P.R. China (e-mail: jiefeng0109@163.com; xiaojian.zhong@stu.xidian.edu.cn; dili@stu.xidian.edu.cn; wsdong@mail.xidian.edu.cn; rhshang@mail.xidian.edu.cn; lchjiao@mail.xidian.edu.cn).}}


\maketitle

\begin{abstract}
Band selection plays a crucial role in hyperspectral image classification by removing redundant and noisy bands and retaining discriminative ones. 
However, most existing deep learning-based methods are aimed at dealing with a specific band selection dataset, and need to retrain parameters for new datasets, which significantly limits their generalizability.
To address this issue, a novel multi-teacher multi-objective meta-learning network (M$^3$BS) is proposed for zero-shot hyperspectral band selection.
In M$^3$BS, a generalizable graph convolution network (GCN) is constructed to generate dataset-agnostic base, and extract compatible meta-knowledge from multiple band selection tasks.
To enhance the ability of meta-knowledge extraction, multiple band selection teachers are introduced to provide diverse high-quality experiences.strategy
Finally, subsequent classification tasks are attached and jointly optimized with multi-teacher band selection tasks through multi-objective meta-learning in an end-to-end trainable way.
Multi-objective meta-learning guarantees to coordinate diverse optimization objectives automatically and adapt to various datasets simultaneously.
Once the optimization is accomplished, the acquired meta-knowledge can be directly transferred to unseen datasets without any retraining or fine-tuning.
Experimental results demonstrate the effectiveness and efficiency of our proposed method on par with state-of-the-art baselines for zero-shot hyperspectral band selection.
\end{abstract}


\ifCLASSOPTIONpeerreview
\begin{center} \bfseries EDICS Category: 3-BBND \end{center}
\fi

\IEEEpeerreviewmaketitle

\section{Introduction}
\label{sec:intro}

\IEEEPARstart{W}{ith} the rapid development of remote sensing technologies, hyperspectral images (HSIs) have become an indispensable embranchment in land cover discrimination.
In comparison with multispectral imaging, hyperspectral imaging measures hundreds of spectral bands for the same spatial area and provides a continuous spectrum with finer wavelength resolution and richer semantic information, which is eligible for image classification~\cite{xu2022robust}, change detection~\cite{qu2024cycle}, and anomaly detection~\cite{lian2024gt}.
Over the past decades, it has been applied in a wide variety of fields, ranging from environmental monitoring, agriculture, mineral and geological exploration, and military applications~\cite{tejasree2024extensive}.

The abundance of spectral bands enables hyperspectral images to perceive and identify land covers accurately.
However, it brings noisy and irrelevant information.
In addition, as the dimensionality increases, the amount of samples needed for training a robust classification model grows exponentially~\cite{Bellman1966Dynamic}.
Furthermore, these hundreds of spectral bands implicitly require significant computing and storage resources.
Therefore, dimensionality reduction is an imperative step for eliminating spectral noise and redundancy, and improving the performance of downstream tasks.

Dimensionality reduction can be fulfilled mainly through two approaches: feature extraction and feature selection~\cite{vaddi2024strategies}.
By applying linear or nonlinear transformations, feature extraction transforms the original feature space into a lower-dimensional one and generates the representations completely different from the original ones.

When feature selection comes to HSIs, it is referred to as an alias for band selection.
As the name suggested, band selection aims at identifying a subset from hundreds of spectral bands to represent the overall spectral information by removing redundant and noisy bands and retaining discriminative ones.
As opposed to feature extraction, band selection excels at retaining original physical information~\cite{Wang2019HyperspectralBS}, making it valuable for real-world applications.

Band selection methods can be roughly divided into filter-based, wrapper-based and embedding-based~\cite{vaddi2024strategies}.
Filter-based methods measure the performance of candidate band subsets by utilizing certain predefined criteria, which are usually independent of the chosen classifiers and thus can be calculated efficiently.
In~\cite{Peng2003FeatureSB}, a representative filter-based method, called minimal-redundancy maximal-relevance (mRMR), was introduced, where the redundancy between any two spectral bands is minimized, and the dependency of bands with class labels is maximized.

Wrapper-based methods treat the classification performance of the chosen classifiers as the evaluation metric of candidate band subsets.
Different from filter-based methods, both evaluation and classification of the selected bands are implemented by the same classifier, which leads to better classification performance and more time-consuming training.
Dynamic classifier selection, abbreviated as DCS, was proposed in~\cite{Cao2019SemiSupervisedHB} as a representative wrapper-based method. DCS utilizes new samples to be classified to select the base classifiers and generates pseudo-labels through edge-preserving filtering.
In~\cite{Ghamisi2015ANF}, fractional-order Darwinian particle swarm optimization (FODPSO) is employed to search for candidate band subsets, and its fitness evaluation is achieved by a support vector machine.
Following this, a particle ranking strategy~\cite{Rashno2022ParticleRA} was proposed to efficiently select features in the multi-objective swarm optimization space.

Embedding-based methods construct an end-to-end framework by combining band selection and classification into a single task, where selection results could be obtained after the training of the chosen classifier.
Hence, these methods not only produce better classification performance than wrapper-based methods but also maintain competitive speed on par with filter-based methods.
In~\cite{Subramanian2002GeneSF}, a SVM based on recursive feature elimination (RFE-SVM) was proposed to maximize the margin by using the weights of the sequential backward selection.
Zhou et al.~\cite{Zhou2018FeatureSM} integrated a genetic algorithm with a SVM to encode and optimize the preliminarily selected feature subsets.
Instead of sorting weight values,~\cite{Kuo2014AKF} proposed a kernel-based SVM to differentiate bands by sorting the magnitude of SVM coefficients, which is more suitable for training SVM classifiers.

Over the past decades, deep learning has made significant advances in solving problems that resisted the best practice of the artificial intelligence community for years~\cite{LeCun2015DeepL}.
Among numerous deep learning techniques suitable for band selection, convolutional neural network (CNN) grows by leaps and bounds due to its excellent spatial representation capacity and powerful nonlinear fitting ability.

During the early stage, CNN-based methods, such as self-improving CNN (SICNN)~\cite{Ghamisi2016ASC}, utilized traditional algorithms like FODPSO to search for candidate subsets.
To evaluate each candidate subset, a two-dimensional CNN is involved in SICNN and needs to be retrained during each iteration.
To mitigate the time consumption caused by retraining, a band selection algorithm based on distance density (DDCNN)~\cite{Zhan2017HyperspectralBS} was proposed.
In DDCNN, a distance density among all the bands is calculated to select band combinations.
Then, a customized one-dimensional CNN is pretrained with original full-band HSIs as an evaluation metric.
Nevertheless, the absence of retraining also brings a decline for the ability of evaluation.
Unlike SICNN and DDCNN, Feng et al.~\cite{Feng2019HyperspectralBS} constructed a novel ternary weight CNN (TWCNN) to indicate whether the corresponding band is selected.
TWCNN combines band selection, feature extraction and classification into a unified end-to-end optimization procedure, comprising a depth-wise convolutional layer, subsequent convolutional layers and fully-connected layers.
Later, a bandwise-independent binary convolution and a novel coarse-to-fine loss are introduced in BHCNN~\cite{Feng2020ConvolutionalNN} to improve the optimization interpretability brought by discrete weights.
Equipped with a novel attention mechanism, an attention-based one-dimensional CNN (ABCNN)~\cite{RibaltaLorenzo2020HyperspectralBS} is coupled with an anomaly detection technique to assign scores to spectral bands and select the most discriminative ones.

In addition to traditional fully supervised settings, unsupervised and semi-supervised methods have also attracted great attentions due to their cheaper annotations.
Cai et al.~\cite{Cai2019BSNetsAE} proposed a novel BS-Nets to formalize band selection as an unsupervised spectral reconstruction task, where the weights are sparsed by l1-regular constraints for band selection of HSIs.
Taking both labeled and unlabeled samples into consideration, Sellami et al.~\cite{Sellami2019HyperspectralIC} proposed a semi-supervised 3D-CNN based on adaptive dimensionality reduction to extract spectral and spatial features for classification.

Apart from convolutional neural networks, a wide variety of advanced techniques in deep learning are energizing band selection by constantly breaking through the theoretical limit of this research field.
Mou et al.~\cite{Mou2021DeepRL} framed unsupervised band selection as a Markov decision process and further exploited reinforcement learning to solve it by training an agent to learn a band-selection policy with two reward schemes.
As in the previous practice, Feng et al.~\cite{Feng2021DeepRL} formalized band selection as a sequential decision-making process in deep reinforcement learning.
For a better measurement on how well a band subset performs, a semi-supervised CNN is constructed as an efficient evaluation criterion.
To make full use of the structural information, Cai et al.~\cite{Cai2020EfficientGC} incorporated graph convolutional layers into a self-representation model for a more robust coefficient matrix to determine an informative band subset.
Similarly, in~\cite{Feng2021DualgraphCN}, insufficient mining towards inter-band correlation can be handled by a dual-graph convolutional network based on a band attention map with a sparse constraint.

It should be noted that, although deep learning-based band selection methods emerge in endlessly and continue to achieve new state-of-the-arts, a common issue still remains unsolved.
Specifically, without considering inherent correlation among different datasets, most existing deep learning-based methods are aimed at handling a specific dataset.
When encountering a new dataset, the model needs to be trained from scratch.
The inherent correlation among different datasets can be referred to as meta-knowledge, that is, the invarient knowledge across different datasets.
Therefore, how to design a dataset-invariant meta-knowledge extractor in zero-shot band selection of HSIs to maintain adaptability over multiple datasets has become a practical but challenging topic. 

In this paper, we propose a novel \textbf{m}ulti-teacher \textbf{m}ulti-objective \textbf{m}eta-learning network (M$^3$BS) for zero-shot hyperspectral \textbf{b}and \textbf{s}election.
Specifically, a spatial-spectral graph is first built to encode spectral bands into a discrete non-Euclidean space, where the vertices are regarded as bands and the edges are determined according to the spatial and spectral relationship among these bands.
On this basis, a generalizable GCN is constructed as a dataset-invariant meta-knowledge extractor which decomposites the parameters to learnable dataset-agnostic bases and dataset-specific coefficients.
After that, subsequent CNN-based classifier is attached.
To guarantee a better generalization ability, several band selection taechers with diversity ensemble strategy is designed to provide extra supervision. 
Finally, an uncertainty-based multi-objective meta-learning procedure is established to jointly optimize band selection and classification in an end-to-end trainable way.
Once the optimization is finished, the dataset-specific coefficients can characterize unseen samples, and the acquired dataset-agnostic bases can be immediately transferred to new datasets without any retraining or fine-tuning. 

The main contributions of this paper are listed as follows:
\begin{enumerate}
    \item
    The generalizable GCN can achieve meta-knowledge extraction that is compatible with various datasets, and make it possible for M$^3$BS to achieve zero-shot band selection of HSIs.
    As the training progresses, inconsistent representations among different datasets will be synchronized by the unified meta-knowledge.
    \item
    Owing to miscellaneous optimization directions, the addition of multifarious experiences from multiple representative and reliable teacher models can maintain generalization and adaptability of M$^3$BS on unseen samples in zero-shot learning, and the generalizable GCN is more prone to accelerate the training procedure and converge in a reasonable time.
    \item
    Using the multi-objective meta-learning end-to-end optimization procedure, diverse optimization objectives can be automatically coordinated without any manual intervention, while multiple band selection tasks from multiple datasets can be simultaneously co-improved.
    As a result, the acquired meta-knowledge can be directly transferred from seen samples to unseen samples without any retraining or fine-tuning.
\end{enumerate}

The remainder of this paper is constituted by the following sections.
With a brief introduction to preliminaries relevant to GCN, multi-objective learning and meta-learning in Section~\ref{sec:bg}, our unified architecture M$^3$BS for zero-shot hyperspectral band selection is described in Section~\ref{sec:method}.
Section~\ref{sec:exp} depicts quantitative statistical experiments and qualitative theoretical analysis with other competitive band selection algorithms.
Section~\ref{sec:concl} ends this paper with a concise conclusion.

\begin{figure*}
    \centering
    \includegraphics[width=\linewidth]{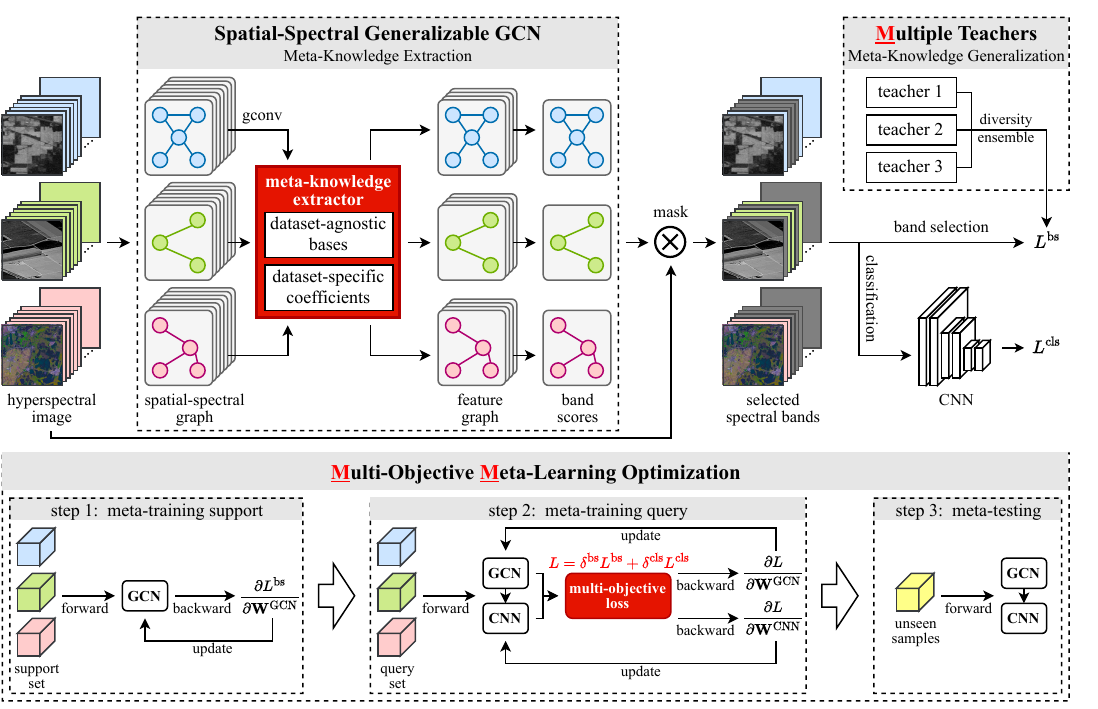}
    \caption{The overall architecture of the proposed M$^3$BS for zero-shot hyperspectral band selection.}
    \label{fig:arch}
\end{figure*}

\section{Background}
\label{sec:bg}

\subsection{Graph Convolutional Network}
\label{subsec:bg-gcn}

GNN~\cite{Scarselli2009TheGN} stands for a series of models capable of extracting and analyzing complicated relationships from a discrete graph.
In contrast to CNN, GNN has shown superior performance and great potential in processing data in a non-Euclidean space.
Since being proposed, GNNs have proven their powerful abstraction ability in various applications and domains, e.g., recommendation system~\cite{Ying2018GraphCN}, traffic forecasting~\cite{Li2017DiffusionCR}, and molecular modeling~\cite{Fout2017ProteinIP}.

By generalizing the idea of convolution operators from a two-dimensional grid space to a discrete graph space, feature representations can be obtained by aggregating the adjacent vertices with a graph convolutional layer.
Let $\graph=\{\vertset,\edgeset,\adjmat\}$ be an undirected acyclic graph, where $\vertset$ and $\edgeset$ are the sets of $\nvert$ vertices and $\nedge$ edges respectively.
Before explaining how a graph convolution operates, the adjacency matrix $\adjmat\in\real^{\nvert\times\nvert}$ is necessary for describing the similarity between two vertices of the graph $\graph$.
Let $\ftrmat\in\real^{\nvert\times\dvert}$ denote an optional matrix associated with feature vectors of each vertex.
With a standard radial basis function~\cite{Buhmann2003RadialBF}, each element in the adjacency matrix can be defined as follows:
\begin{equation}
    \adjmat_{\ivert,\jvert} = \exp{\left(-\frac{\Vert \ftrmat_\ivert-\ftrmat_\jvert \Vert_2^2}{\sigma^2}\right)}
    \label{eqn:rbf}
\end{equation}
where $\sigma$ is the width constant, $\ftrmat_\ivert$ and $\ftrmat_\jvert$ are two rows of the feature matrix, corresponding to the feature vector of the $\ivert$-th and $\jvert$-th vertex, respectively.
For a better generalization ability~\cite{Feige2019SpectralGT}, the symmetric normalized Laplacian matrix $\laplamat$ is expressed as:
\begin{equation}
    \laplamat = \idmat - \degmat^{-\half} \adjmat \degmat^{-\half}
    \label{eqn:lapla}
\end{equation}
where $\degmat$ is the degree matrix of the adjacency matrix, that is, $\degmat_{\ivert,\jvert} = \sum_\jvert\adjmat_{\ivert,\jvert}$.
The Laplacian matrix can be decomposited by a spectral decomposition $\laplamat = \mathbf{U} \mathbf{\Lambda} \mathbf{U}^\top$.
Given two functions $f(\cdot)$ and $g(\cdot)$, the graph convolution operator $\gconv$ can be defined as Eq.~\eqref{eqn:gconv} by regarding $g_\theta = \mathbf{U}^\top g$ as the convolutional kernel:
\begin{equation}
    f \gconv g
    = \mathbf{U} \{ (\mathbf{U}^\top f) \cdot (\mathbf{U}^\top g) \}
    = \mathbf{U} g_\theta \mathbf{U}^\top f
\label{eqn:gconv}
\end{equation}
This formula imitates the way a traditional convolution $\conv$ is defined, i.e., transforming the graph into the spectral domain by a Fourier transform $\mathbf{U}^\top \cdot$, performing a matrix multiplication, and converting back to the spatial domain by an inverse Fourier transform $\mathbf{U} \cdot$.

As one of the most famous and practical forms of GCNs, the propagation rule was proposed by Kipf et al.~\cite{Kipf2016SemiSupervisedCW}, which is also used in this paper. It diminishes the computational cost by introducing a constrained Chebyshev polynomial significantly:
\begin{equation}
    \mathbf{Y} = \actfun( \tilde\degmat^{-\half} \tilde\adjmat \tilde\degmat^{-\half} \ftrmat \wgtmat + \biasvec )
\end{equation}
where $\tilde\adjmat = \adjmat + \idmat$ and $\tilde\degmat_{\ivert,\jvert} = \sum_\jvert\tilde\adjmat_{\ivert,\jvert}$ are renormalized from $\adjmat$ and $\degmat$ for a better training stability,
$\actfun(\cdot)$ is a custom activation function,
$\wgtmat \in \real^{\dvert \times \dvert^\prime}$ and $\biasvec \in \real^{\dvert^\prime}$ are weights and bias that serve as learnable parameters.
This graph convolutional layer takes a feature matrix $\wgtmat$ with $\nvert$ $\dvert$-dimensional vectors as the input, and outputs another feature matrix $\mathbf{Y}$ with the same number of $\dvert^\prime$-dimensional vectors.

\subsection{Multi-Objective Learning}
\label{subsec:bg-mtl}

Multi-objective learning (MTL)~\cite{Caruana1993MultitaskLA} aims to handle multiple different tasks with a single neural network, in which parameters are divided into shared ones and independent ones~\cite{Hou2020LargeScaleEM}.
Compared to conventional single-task learning, MTL does better in reducing the amount of parameters by efficiently leveraging common knowledge and similarities among associated tasks with shared layers.
For example, MTL-enhanced feature extraction layers inside a CNN are compatible with various computer vision tasks such as classification, detection and segmentation~\cite{He2017MaskR}.
Furthermore, inductive biases from multiple tasks during training enhance the robustness of the model in disguise, making it less susceptible to noise samples.
The generality of MTL makes it suitable for introducing auxiliary tasks to improve the accuracy of the main task, e.g., classifying gender and pose for facial landmark detection~\cite{Zhang2014FacialLD}, discriminating speakers for voice trigger detection~\cite{Sigtia2020MultiTaskLF}.
In addition to designing model structures to support multiple inputs and outputs, it is also crucial for MTL to consider how to balance loss functions from different tasks.
In this paper, a single-input-multi-output MTL framework with hard parameter sharing is adopted to determine the loss weights of both band selection and classification tasks automatically.

\subsection{Meta-Learning}
\label{subsec:bg-meta}

Since Donald Maudsley coined the meta-learning in 1979 ~\cite{Maudsley1979Theory} to describe internalized perception, inquiry, learning and growth, it has increasingly attracted attentions and is considered as a golden key to achieve general artificial intelligence.
The idea behind meta-learning is the so-called ``learn-to-learn'', that is, taking advantages of knowledge and experiences from a variety of existing tasks to acquire the learning ability and quickly generalize to new tasks.
Different from the well-known machine learning pipeline, the minimum granularity of meta-learning is no longer a single sample, but a complete task composed of multiple samples.
Besides, instead of a function that maps from images to labels, the goal of meta-learning is a meta-function that generates functions for specific tasks.
Regarded as a sub-field of transfer learning, meta-learning transfers knowledge from the source domain of seen tasks to the target domain of unseen tasks.
Existing approaches towards meta-learning can be roughly divided into three categories: learning weight initializations, training meta-models that generates model parameters, as well as designing transferable optimizers~\cite{Finn2017ModelAgnosticMF}.
Promising applications of meta-learning have been presented in various areas spanning few-shot learning~\cite{Shaban2017OneShotLF}, meta reinforcement learning~\cite{Young2018MetatraceAO} and neural architecture search~\cite{Shaw2018MetaAS}.
In this paper, zero-shot optimization for band selection is formalized as a two-stage meta-learning procedure.

\begin{figure*}
    \centering
    \includegraphics[width=\linewidth]{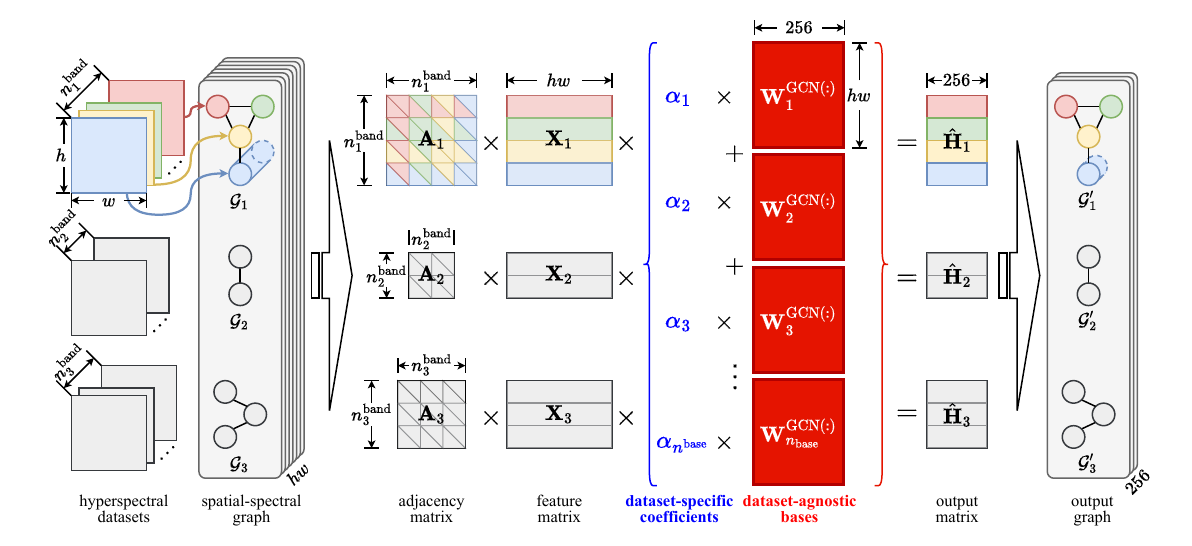}
    \caption{The first graph convolutional layer inside the generalizable GCN, comprising consecutive matrix multiplications. The weight matrix is parameterized as dataset-agnostic bases and dataset-specific coefficients. These bases are shared by different datasets, while these coefficients, which are specific to each dataset, are computed by a simple MLP network.}
    \label{fig:gcn-gconv1}
\end{figure*}

\section{Methodology}
\label{sec:method}

The overall architecture of the proposed M$^3$BS for zero-shot hyperspectral band selection is illustrated in Fig.~\ref{fig:arch}.
Specifically, the generalizable GCN with spatial-spectral graphs is constructed for dataset-agnostic band selection in Sec.~\ref{subsec:method-gcn}, followed by subsequent classification in Sec.~\ref{subsec:method-cnn}.
In Sec.~\ref{subsec:method-bs}, auxiliary supervision from multiple band selection teachers is introduced with a diversity ensemble strategy.
Sec.~\ref{subsec:method-optim} focuses on zero-shot optimization, where all the modules are integrated into an unified uncertainty-based multi-objective meta-learning framework to jointly optimize band selection and classification tasks.

\subsection{Spatial-Spectral Generalizable GCN for Band Selection}
\label{subsec:method-gcn}

Aiming at zero-shot band selection, we construct a generalizable GCN with spatial-spectral graphs to calculate an importance score for each band.
It is served as a dataset-invariant meta-knowledge extractor to decomposite the parameters to learnable dataset-agnostic bases and dataset-specific coefficients.
Thus, it can synchronize inconsistent representations among different datasets, and finally learn ``how to perform band selection'' and acheive zero-shot inference on any new dataset.

\subsubsection{Spatial-Spectral Graph}

In contrast to the traditional grid structure, the graph structure can capture complex and irregular associations among individuals, which makes it particularly eligible for modeling correlation among spectral bands in HSIs. 
In this way, the bands and their spatial and spectral relationships are regarded as vertices and edges, respectively.

Suppose there are $\ndataset$ different HSI datasets $\{ \dataset_1, \dots, \dataset_\ndataset \}$, each of which consists of a different number of bands $\nband$ but shares a common patch size $\height \times \width$.
As for the $\idataset$-th dataset $\dataset_\idataset$, the proposed spatial-spectral band graph is represented as $\graph = \{\vertset, \edgeset, \adjmat\}$.
$\vertset$ is the set of vertices corresponding to spectral bands of HSIs, and $\vertelm_{\iband} \in \vertset$ denotes the $\iband$-th spectral band.  
$\edgeset$ is the set of edges indicating the relationships among all the spectral bands, and $\edgeelm_{\iband,\jband} = (\vertelm_{\iband}, \vertelm_{\jband}) \in \edgeset$ denotes the correlation between the $\iband$-th and $\jband$-th bands.
By flattening all the pixels in a single patch, the resulting feature vector $\ftrvec_{\iband} \in \real^{\dband}$ corresponds to the $\iband$-th band $\vertelm_{\iband}$, and the feature matrix $\ftrmat \in \real^{\nband \times \dband}$ is formed by stacking together all these feature vectors.
As illustrated in Fig.~\ref{fig:gcn-gconv1}, the ``blue'' row of the feature matrix corresponds to the ``blue'' spectral band, the ``green'' row corresponds to the ``green'' spectral band, and so on.

The adjacency matrix, denoted as $\adjmat \in \real^{\nband \times \nband}$ is utilized for describing the similarity between two vertices.
We exploit both spatial and spectral information with a threshold for truncation $\adjthres$ to calculate the adjacency matrix as follows:
\begin{equation}
    \adjmat_{\iband,\jband} =
    \begin{cases}
        \adjmat_{\iband,\jband}^\text{spa} + \adjmat_{\iband,\jband}^\text{spec},&\text{if } \adjmat_{\iband,\jband}^\text{spa} + \adjmat_{\iband,\jband}^\text{spec} \ge \adjthres \\
        0,&\text{otherwise}
    \end{cases}
    \label{eqn:adj}
\end{equation}
\begin{equation}
    \adjmat_{\iband,\jband}^\text{spa} =
    \begin{cases}
        \exp \left( - \cfrac{1}{\nband} \vert \iband-\jband \vert \right),&\text{if } \iband \neq \jband \\
        0,&\text{otherwise}
    \end{cases}
    \label{eqn:adj-spa}
\end{equation}
\begin{equation}
    \adjmat_{\iband,\jband}^\text{spec} =
    \begin{cases}
        \exp \left( - \cfrac{1}{\dband} \Vert \ftrvec_{\iband} - \ftrvec_{\jband} \Vert_2 \right),&\text{if } \iband \neq \jband \\
        0,&\text{otherwise}
    \end{cases}
    \label{eqn:adj-spec}
\end{equation}
where $\adjmat^\text{spa}$ and $\adjmat^\text{spec}$ are critical to characterizing the similarity among spectral bands from different aspects.
Concretely, $\adjmat_{\iband,\jband}^\text{spa}$ indicates the $1$-dimensional spatial distance between the $\iband$-th and $\jband$-th bands. The closer the two bands are, the closer the value is to $1$.
The spatial distance reflects the fact that the intensity of adjacent bands is tend to be close due to their similar light reflectance.
$\adjmat_{\iband,\jband}^\text{spec}$ indicates the $l^2$-norm of the difference of the feature vector $\ftrvec_{\iband}$ and $\ftrvec_{\jband}$, in the sense of the $\dband$-dimensional Euclidean distance.
As shown in Eq.~\eqref{eqn:adj-spa} and ~\eqref{eqn:adj-spec}, the Gaussian kernel function is utilized to constrain the numeric range of the two similarity matrices $\adjmat^\text{spa}$ and $\adjmat^\text{spec}$ between $0$ and $1$.
Besides, as introduced in Section~\ref{sec:bg}, we still apply renormalization to the adjacency matrix $\widetilde{\adjmat} = \adjmat + \idmat$, thereby enhancing stability in the training process.

\subsubsection{Dataset-Agnostic Generalizable GCN}

Meta-knowledge can be understood as generic knowledge acquired from continuous adaptation to different tasks.
For meta-knowledge extraction, we propose a more effective representation of learnable parameters, which decomposites these parameters into dataset-agnostic bases and dataset-specific coefficients.
These dataset-agnostic bases act as unified meta-knowledge to synchronize inconsistent representations among different datasets, while dataset-specific coefficients can effectively deal with unseen samples due to their dynamically generated characteristics.

Based on the abstraction of the spatial-spectral band graph, a spatial-spectral generalizable GCN is constructed for band selection in M$^3$BS, comprising two graph convolutional layers with batch normalization (BN) as shown in Eq.~\eqref{eqn:gcn-gconv1} and \eqref{eqn:gcn-gconv2}:
\begin{equation}
    \hidmat = \relu( \bn( \widetilde{\degmat}^{-\half} \widetilde{\adjmat} \widetilde{\degmat}^{-\half} \ftrmat \wgtmat^\text{GCN(1)} ) )
    \label{eqn:gcn-gconv1}
\end{equation}
\begin{equation}
    \scrvec = \sigmoid( \bn( \widetilde{\degmat}^{-\half} \widetilde{\adjmat} \widetilde{\degmat}^{-\half} \hidmat \wgtmat^\text{GCN(2)} ) )
    \label{eqn:gcn-gconv2}
\end{equation}
where $\widetilde{\adjmat}$ and $\widetilde{\degmat}$ are the renormalized forms of the adjacency matrix and the degree matrix respectively, $\hidmat \in \real^{\nband \times \dhid}$ denotes the output feature matrix of the first graph convolutional layer, $\scrvec \in [0, 1]^{\nband}$ indicates the importance score of each band for determining which subset should be selected, $\wgtmat^\text{GCN(1)} \in \real^{\dband \times \dhid}, \wgtmat^\text{GCN(2)} \in \real^{\dhid}$ are learnable parameters, which will be updated by gradient descent during training.
The output dimensionality of each vertex in the first layer is $\dhid$.
The second layer is normalized by a Sigmoid function $\sigmoid(\cdot)$ to ensure that the score of each band falls within $[ 0, 1 ]$.

Instread of static forward propagation, we parameterize the above-mentioned graph convolutional kernels $\wgtmat^\text{GCN(1)}$ and $\wgtmat^\text{GCN(2)}$ as linear combinations of $\nbase$ bases:
\begin{equation}
    \wgtmat^\text{GCN(:)} = \alpha_1\wgtmat_1^\text{GCN(:)} + \dots + \alpha_{n^\text{base}}\wgtmat_\nbase^\text{GCN(:)}
    \label{eqn:gcn-condconv}
\end{equation}
where $\wgtmat_1^\text{GCN(:)} \dots \wgtmat_\nbase^\text{GCN(:)}$ are relatively orthogonal bases decomposited from the original parameter, $\alpha_1 \cdots \alpha_\nbase$ are linear combination coefficients from an average pooling layer, a fully-connected layer and a sigmoid activation function:
\begin{equation}
    \alpha_1 \dots \alpha_\nbase = \sigmoid( \frac{1}{\nband} \mathbf{1}^\top \ftrmat \wgtmat^\text{FC} )
    \label{eqn:gcn-condconv-alpha}
\end{equation}
where $\frac{1}{\nband} \mathbf{1}^\top \in \real^{1 \times \nband}$ calculates the mean along each column of the feature matrix $\ftrmat$,  $\wgtmat^\text{FC} \in \real^{\height\width \times \nbase}$ maps the pooled feature matrix into $\nbase$ scalars corresponding to $\nbase$ parameter bases.
Meta-knowledge is explicitly represented as dataset-agnostic bases, while dataset-specific coefficients can be used to characterize unique attributes of different datasets.

According to the extracted score vector $\scrvec$ for all the bands, it is available to determine which subset should be selected.
Generally, the number of selected bands is fixed as a hyperparameter $\nsband$ by users in advance ($\nsband < \nband$)~\cite{Feng2020ConvolutionalNN}.
To make it available to attach the following classification network, a binarization operator is devised to convert scores into a band mask:
\begin{equation}
    \scrthres = \left\{ \scrvec_{o_1}, \scrvec_{o_2}, \cdots, \scrvec_{o_\nband} \right\}_{\nsband}
    \label{eqn:threshold}
\end{equation}
\begin{equation}
    \mskvec
    = \bin(\scrvec, \scrthres)
    =
    \begin{cases}
        0,&\text{if } \scrvec_{\iband} \geq \scrthres \\
        1,&\text{otherwise}
    \end{cases}
    \label{eqn:binarization}
\end{equation}

Specifically, the scores of all the bands are firstly sorted in the descending order $o_1, o_2, \cdots, o_{\nband}$, and the threshold $\scrthres$ is taken to be the $\nsband$-th score.
Then, a binarization operator $\bin(\cdot)$ is applied to extract a band mask $\mskvec \in \{0,1\}^{\nband}$ by providing a conditional comparative constraint, where the corresponding value is set to $0$ when the score is smaller than or equals to the threshold.
Finally, this band mask is multiplied to the HSIs as the input of the classifier.

\subsection{CNN-based Image Classifier}
\label{subsec:method-cnn}

\begin{table}
    \caption{The detailed structure of the CNN-based image classifier.}
    \begin{center}
    \begin{tabular}{c|c|c|c}
    \toprule[1.5pt]
    Stage & Operator & Resolution & \#Channels \\
    $\istage$ & $\hat{\operator_\istage}$ & $\hat{\height}_\istage \times \hat{\width}_\istage$ & $\hat{\nchannel}_\istage$ \\
    \midrule[1.0pt]
    0 & -                               & $33 \times 33$ & $\nsband$  \\
    1 & ConvBNReLU, 5$\times$5 w/ pad=2 & $33 \times 33$ & $64$   \\
    1 & MaxPool, 2$\times$2             & $16 \times 16$ & $64$   \\
    2 & ConvBNReLU, 5$\times$5 w/ pad=2 & $16 \times 16$ & $128$  \\
    2 & MaxPool, 2$\times$2             & $8 \times 8$   & $128$  \\
    3 & ConvBNReLU, 5$\times$5 w/ pad=2 & $8 \times 8$   & $256$  \\
    3 & MaxPool, 2$\times$2             & $4 \times 4$   & $256$  \\
    4 & ConvBNReLU, 5$\times$5 w/ pad=2 & $4 \times 4$   & $512$  \\
    4 & MaxPool, 2$\times$2             & $2 \times 2$   & $512$  \\
    5 & ConvBNReLU, 5$\times$5 w/ pad=2 & $2 \times 2$   & $1024$ \\
    5 & MaxPool, 2$\times$2             & $1 \times 1$   & $1024$ \\
    6 & Flatten                         & -              & $1024$ \\
    6 & Dropout \& FC \& Softmax        & -              & $\nclass$  \\
    \bottomrule[1.5pt]
    \end{tabular}
    \end{center}
    \label{tab:cnn}
\end{table}

To embed an auxiliary classification network to evaluate band combinations, we construct a 2-dimensional CNN to extract spatial features and further output a probability distribution for classification.  

Suppose that there are $\nclass$ categories in the $\idataset$-th HSI dataset.
The detailed network structure of the CNN for image classification is illustrated in Table~\ref{tab:cnn}, consisting of five stages of spatial convolutions and one stage of dense connection.
With the hyperspectral image patch $\patch \in \real^{\nband \times \height \times \width}$ as the input, the probability distribution $\prbvec \in [0, 1]^{\nclass}$ can be predicted by the $k$-th CNN with parameters $\wgtmat^\text{CNN}_\idataset$.

Following the settings of embedding-based band selection methods, the classification loss for a single batch is defined as a multi-class cross-entropy loss as shown in Eq.~\eqref{eqn:loss-cls}.
$\clsgtvec_{\ipatch} \in \{1, 2, \cdots, \nclass\}$ denotes which category the center pixel of the $i$-th patch actually belongs to, and $\mathbf{1}\{\cdot\}$ is a vectorized indicator function to perform one-hot encoding over scalars.
\begin{equation}
    \loss^\text{cls} = -\frac{1}{\batchsize} \sum\limits_{\ipatch=1}^{\batchsize} \mathbf{1}\{ \clsgtvec_{\ipatch} \} \log\prbvec_{\ipatch}
    \label{eqn:loss-cls}
\end{equation}

\subsection{Multi-Teacher Diversity Ensemble}
\label{subsec:method-bs}

As a typical preprocessing procedure for downstream tasks, band selection does not involve any explicit ground truths like other fully supervised tasks, such as image classification.
Thus, it is usually formalized as an unsupervised task~\cite{Cai2019BSNetsAE} or evaluated by the performance of an auxiliary classifier~\cite{Cao2019SemiSupervisedHB}.
These band selection methods either produce poor performance or consume numerous training time.
Moreover, they fail to generalize to unseen samples, which makes it challenging to acheive zero-shot hyperspectral band selection.

In addition to the classification loss in Sec.~\ref{subsec:method-cnn}, we exploit diverse high-quality experiences from multiple band selection teachers, generating an auxiliary loss function.
This extra supervision not only facilitates generalization and adaptability for the previous meta-knowledge, but also accelerates the convergence procedure and alleviates the training time consumption.

Specifically, multiple representative and reliable teachers from filter-based, wrapper-based and embedding-based methods respectively, are pretrained to prepare for high-quality experiences in advance.
Considering how to make a balance among multiple teachers, we formulate a diversity ensemble strategy to choose the spectral bands according to their popularity.
After the band subsets are selected by these teachers respectively, a counting function $\text{cnt}(i)$ can be defined to denote the number of received votes of the $\iband$-th band.
As stated in Equation~\ref{eqn:divens-set} and~\ref{eqn:divens}, we can obtain a band sequence and take the top $\nsband$ bands as the ground truth by sorting these bands by the number of received votes in the descending order:
\begin{equation}
    \mathbb{S} = \left\{ \underbrace{i_1, \dots}_{\text{cnt}(i) = 3} \underbrace{i_7, i_8, \dots \dots}_{\text{cnt}(i) = 2} \underbrace{i_{22}, i_{23}, \dots \dots \dots}_{\text{cnt}(i) = 1} \right\} _{1, \dots, \nsband}
    \label{eqn:divens-set}
\end{equation}
\begin{equation}
    \bsgtvec_i =
    \begin{cases}
        1,&\text{if } i \in \mathbb{S} \\
        0,&\text{otherwise}
    \end{cases}
    \label{eqn:divens}
\end{equation}
where $\bsgtvec$ denotes the label for directly supervising the band selection task.
For the sake of fairness, for those bands with the same number of votes, we randomly pick a subset with an appropriate number of bands.
With this ensemble strategy, multiple teachers can be integrated into a more powerful teacher.
Thanks to the various optimizing directions from different teachers, the lack of generalization can be effectively alleviated, and this kind of diversity is beneficial to generalizing better to unseen samples in zero-shot learning.

The band selection loss for the entire batch is defined as a multi-label binary cross-entropy loss of the continuous importance scores $\scrvec \in [0, 1]^{\nband}$ from the GCN and the discrete selected bands $\bsgtvec \in \{0, 1\}^{\nband}$ from the integrated teacher:
\begin{equation}
    \loss^\text{bs} = \bsgtvec \log\scrvec + (\mathbf{1} - \bsgtvec) \log( \mathbf{1} - \scrvec )
    \label{eqn:loss-bs}
\end{equation}

\subsection{Uncertainty-based Multi-Objective Learning}
\label{subsec:method-mtl}

Both the classification loss in Sec.~\ref{subsec:method-cnn} and the selection loss in Sec.~\ref{subsec:method-bs} are beneficial for the expected band selection task.
It inspires us to simultaneously optimize them with an integrated loss function.
Intuitively, combining multiple losses could be simply accomplished by applying a weighted sum and constituting the overall loss function $\loss$ as follows:
\begin{equation}
    \loss = \weight^\text{bs} \loss^{\text{bs}} + \weight^\text{cls} \loss^{\text{cls}}
    \label{eqn:loss}
\end{equation}
where $\weight^\text{bs} \in [0,1]$ and $\weight^\text{cls} \in [0,1]$ are relative weights for the band selection loss and the classification loss respectively.
In $\ndataset$ different HSI datasets, there are $2\ndataset$ extra hyperparameters needed to be tuned and denoted as $\{ \weight_1^\text{cls}, \dots, \weight_\ndataset^\text{cls}, \weight_1^\text{bs}, \dots, \weight_\ndataset^\text{bs} \}$.
Acutually, the performance of the band selection task is heavily sensitive to the propotion of these weights~\cite{Liu2018EndToEndML}.
However, it is difficult to tune these hyperparameters manually.

It is more ideal to allow the loss weights to be updated together with network parameters during the training process.
Inspried by~\cite{Kendall2017MultitaskLU}, a multi-objective learning approach is proposed, which weighs multiple loss functions by considering the uncertainty of each task.
\cite{Kendall2017MultitaskLU} stated that the accidential error between model outputs and sample labels can be modeled as a homoscedastic uncertainty, which allows us to decompose the multi-objective loss into the product of multiple likelihoods.
By estimating these two tasks with a Sigmoid likelihood $\sigmoid(\cdot)$ and a Softmax likelihood $\softmax(\cdot)$ respectively, an uncertainty-based minimization objective can be defined as follows:
\begin{align}
\label{eqn:loss-uncertainty}
    \loss
    &= -\log \sigma(\scrvec_\idataset;\ \wgtmat^\text{GCN}, \weight^\text{bs}) \nonumber\\
    &\qquad\qquad \cdot \softmax(\prbvec_\idataset;\ \wgtmat^\text{GCN}, \wgtmat_\idataset^\text{CNN}, \weight^\text{cls}) \nonumber\\
    &\propto \weight^\text{bs} \loss^\text{bs} + \weight^\text{cls} \loss^\text{cls} + \log\sqrt{\frac{1}{\weight^\text{bs}}} + \log\sqrt{\frac{1}{\weight^\text{cls}}}
\end{align}
where $\weight^\text{bs}$ and $\weight^\text{cls}$ are two weights corresponding to the above-mentioned two objectives.
As $\weight^\text{bs}$ or $\weight^\text{cls}$ decreases, $1/(\weight^\text{bs})^2$ or $1/(\weight^\text{cls})^2$ increases, which brings a greater influence to the corresponding loss function.
These relative weights are also discouraged from decreasing excessively by the last regularization term $\log\sqrt{1 / {\weight^\text{bs}}}$ and $\log\sqrt{1 / {\weight^\text{cls}}}$.

By applying Eq.~\eqref{eqn:loss-uncertainty} rather than Eq.~\eqref{eqn:loss} to optimize multiple objectives, massive hyperparameter tuning costs can be avoided since the weights are determined automatically by gradient descent without any manual intervention, and different training stages can be adapted since the weights are updated dynamically as training progresses.

\subsection{Multi-Objective Meta-Learning for Zero-Shot Optimization}
\label{subsec:method-optim}

\begin{figure}
    \centering
    \includegraphics{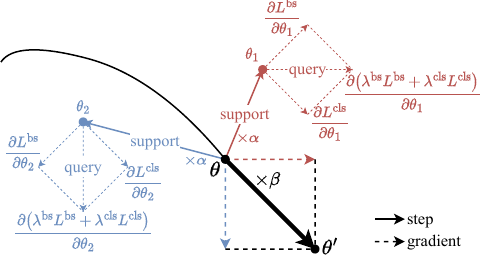}
    \caption{The multi-objective multi-objective meta-learning procedure for zero-shot optimization, where different colors correspond to different HSI datasets.}
    \label{fig:optim}
\end{figure}

When it comes to zero-shot inference, we try to figure out how to perform hyperspectral band selection on a new dataset without any of its samples during training.
To achieve zero-shot optimization, we construct a multi-objective meta-learning procedure to formalize an intuitive trial-and-error process.
It makes our framework not only suitable for both the band selection and classification objectives, but also applicable to various datasets with different characteristics, thereby ensuring maximum compatibility.

Following how previous works preprocess data for meta-learning~\cite{Shaban2017OneShotLF, Young2018MetatraceAO, Shaw2018MetaAS}, we divide HSI datasets into meta-training and meta-testing tasks.
Our goal is to acheive meta-knowledge extraction from meta-training tasks and apply them to meta-testing tasks. 
Specifically, let HSI datasets $\dataset_1^\text{train}, \dots, \dataset_\ndataset^\text{train}$ be different meta-training tasks for training, $\dataset^\text{test}$ be the meta-testing task for zero-shot inference.
What we want to achieve is to directly inference on $\dataset^\text{test}$ after training on $\dataset_1^\text{train}, \dots, \dataset_\ndataset^\text{train}$, and none of samples in $\dataset^\text{test}$ are provided during the training phrase.
Besides, each meta-training task (i.e. dataset) $\dataset_\idataset^\text{train}$ is further divided into a support set $\dataset_\idataset^\text{spt}$ and a query set $\dataset_\idataset^\text{qry}$ according to a predefined percentage.

The overall procedure of M$^3$BS is described elaborately in Alg.~\ref{alg}.
Inspired by model-agnostic meta-learning~\cite{Finn2017ModelAgnosticMF}, the whole meta-training process can be viewed as a trial-and-error process in Fig.~\ref{fig:optim}, where the support stage makes an attempt and the query stage verifies the effectiveness of this attempt.
Significantly different from~\cite{Finn2017ModelAgnosticMF}, instead of applying the same loss function for both stages, we optimize the multi-objective loss in Sec.~\ref{subsec:method-mtl} in the query stage, thereby utilizing the stronger supervision from classification to more effectively verify whether the attempt made in the support stage is reasonable.
In the support stage, only the band selection loss $\loss^\text{bs}$ is involved for a tentative update step for the temporary parameters $\wgtmat^\text{GCN}_\idataset$.
In the query stage, the multi-objective loss $\loss$ is used to verify whether the previous step is beneficial for classification or not.
For verification, only the derivatives of the loss with respect to the parameters $\partial{\loss}/\partial{\wgtmat^\text{GCN}_\idataset}$ are calculated, and the parameters $\wgtmat^\text{GCN}_\idataset$ are not actually updated.
After iterating over all $\ndataset$ meta-training tasks, all these derivatives are translated to the original parameters $\wgtmat^\text{GCN}\vert_\iepoch$ for gradient descent, leading to a more generalizable version $\wgtmat^\text{GCN}\vert_{\iepoch+1}$.

During meta-testing, discriminative and informative band subsets can be obtained by applying the compatible meta-knowledge extracted by the generalizable GCN to perform zero-shot inference.

\begin{algorithm}[t]
\caption{M$^3$BS}
\begin{algorithmic}[1]
\small{
    \REQUIRE
    $\dataset_1^\text{train}, \dots, \dataset_\ndataset^\text{train}, \dataset^\text{test}$: meta-training and meta-testing tasks;
    $\lr$: learning rate;
    $\mlr$: meta-learning rate;
    $\nepoch$: number of epochs.
    \ENSURE
    $\wgtmat^\text{GCN(:)}, \wgtmat^\text{CNN}_1, \dots, \wgtmat^\text{CNN}_\ndataset$: well-train parameters.
    \STATE Initialize relative weights: $\weight^\text{bs}\vert_{t=1} \leftarrow 1, \weight^\text{cls}\vert_{t=1} \leftarrow 1$
    \FOR {epoch $\iepoch = 1, \dots, \nepoch$}
        \FOR {each meta-training task $\dataset^\text{train}_\idataset$ in $\{\dataset^\text{train}_1, \dots, \dataset^\text{train}_\ndataset\}$}
            \STATE Split samples for support and query: $\dataset_\idataset \rightarrow \{ \dataset^\text{spt}_\idataset, \dataset^\text{qry}_\idataset \}$ 
            \STATE Temporarily duplicate parameters: $\wgtmat^\text{GCN}_\idataset \leftarrow \wgtmat^\text{GCN}$ 
            \FOR {each batch in $\dataset^\text{spt}_\idataset$}
                \STATE Flatten all pixels into a feature matrix $\ftrmat$
                \STATE Characterize an adjacency matrix $\adjmat$ by \eqref{eqn:adj}-\eqref{eqn:adj-spec}
                \STATE Build a spatial-spectral graph $\graph = \{ \vertset, \edgeset, \adjmat \}$
                \STATE Assign importance scores $\scrvec$ with $\wgtmat^\text{GCN}$ by \eqref{eqn:gcn-gconv1}-\eqref{eqn:gcn-condconv-alpha}
                \STATE Vote a band subset $\bsgtvec$ from teachers by \eqref{eqn:divens-set}-\eqref{eqn:divens}
                \STATE Measure the band selection loss $\loss^\text{bs}$ by \eqref{eqn:loss-bs}
                \STATE Backpropagate gradients to obtain $\partial{\loss^\text{bs}}/\partial{\wgtmat_\idataset^\text{GCN}}$
                \STATE Update $\wgtmat^\text{GCN}_\idataset$ with gradient descent:\\
                $\wgtmat^\text{GCN}_\idataset\vert_{\iepoch+1} = \wgtmat^\text{GCN}_\idataset\vert_\iepoch + \lr \cdot \partial{\loss^\text{bs}}/\partial{\wgtmat^\text{GCN}_\idataset\vert_\iepoch}$
            \ENDFOR
            \FOR {each batch in $\dataset^\text{qry}_\idataset$}
                \STATE Obtain $\scrvec$ and $\loss^\text{bs}$ by following the same steps as above
                \STATE Binarize $\scrvec$ into a band mask $\mskvec$ by \eqref{eqn:threshold}-\eqref{eqn:binarization}
                \FOR {each patch $\patch_\ipatch$ and category $\clsgtvec_\ipatch$ in this batch}
                    \STATE Mask $\patch_\ipatch$ with $\mskvec$ by multiplication: $\patch_\ipatch \leftarrow \patch_\ipatch \cdot \mskvec$
                    \STATE Classify $\patch_\ipatch$ into a probability vector $\prbvec_\ipatch$ with $\wgtmat^\text{CNN}_\idataset$
                \ENDFOR
                \STATE Measure the classification loss $\loss^\text{cls}$ by \eqref{eqn:loss-cls}
                \STATE Measure the multi-objective loss $\loss$ by \eqref{eqn:loss-uncertainty}
                \STATE Backpropagate gradients to obtain\\$\partial{\loss}/\partial\weight^\text{bs}$, $\partial{\loss}/\partial\weight^\text{cls}$, $\partial{\loss}/\partial{\wgtmat^\text{GCN}_\idataset}$ and $\partial{\loss}/\partial{\wgtmat^\text{CNN}_\idataset}$
                \STATE Accumulate $\partial{\loss}/\partial{\wgtmat^\text{GCN}_\idataset}$ for $\wgtmat_\idataset^\text{GCN}$
                \STATE Update $\weight^\text{bs}$, $\weight^\text{cls}$ and $\wgtmat_\idataset^\text{CNN}$ with gradient descent:\\
                $\weight^\text{bs}\vert_{\iepoch+1} = \weight^\text{bs}\vert_\iepoch + \lr \cdot \partial{\loss}/\partial{\weight^\text{bs}\vert_\iepoch}$\\
                $\weight^\text{cls}\vert_{\iepoch+1} = \weight^\text{cls}\vert_\iepoch + \lr \cdot \partial{\loss}/\partial{\weight^\text{cls}\vert_\iepoch}$\\
                $\wgtmat^\text{CNN}_\idataset\vert_{\iepoch+1} = \wgtmat^\text{CNN}_\idataset\vert_\iepoch + \lr \cdot \partial{\loss}/\partial{\wgtmat^\text{CNN}_\idataset\vert_\iepoch}$
            \ENDFOR
            \STATE Average $\partial{\loss}/\partial{\wgtmat^\text{GCN}_\idataset}$ over all batches for $\wgtmat_\idataset^\text{GCN}$
            \STATE Summarize $\partial{\loss}/\partial{\wgtmat^\text{GCN}_\idataset}$ to $\partial{\loss}/\partial{\wgtmat^\text{GCN}}$ for $\wgtmat^\text{GCN}$:\\
            $\partial{\loss}/\partial{\wgtmat^\text{GCN}} \leftarrow \partial{\loss}/\partial{\wgtmat^\text{GCN}} + \partial{\loss}/\partial{\wgtmat^\text{GCN}_\idataset}$
        \ENDFOR
        \STATE Update $\wgtmat^\text{GCN}$ with gradient descent for meta-learning:\\
        $\wgtmat^\text{GCN}\vert_{\iepoch+1} = \wgtmat^\text{GCN}\vert_\iepoch + \mlr \cdot \partial{\loss}/\partial{\wgtmat^\text{GCN}}$
    \ENDFOR
}
\end{algorithmic}
\label{alg}
\end{algorithm}

\section{Experimental Results and Analysis}
\label{sec:exp}

In this section, quantitative statistical experiments and qualitative theoretical analysis are conducted over three commonly used hyperspectral datasets to validate the effectiveness and efficiency of our method.
Detailed descriptions of these datasets and experimental settings are provided in Sec.~\ref{subsec:exp-desc} and Sec.~\ref{subsec:exp-setting}, respectively.

\subsection{Dataset Description}
\label{subsec:exp-desc}

\begin{table}
    \caption{Categories and Respective Numbers of Samples}
    \label{tab:categories}
    \begin{center}
    \begin{tabular}{c|c|c|r}
    \toprule[1.5pt]
    Dataset & No. & Category & \#Samples \\
    \midrule[1.0pt]
    \multirow{17}*{\makecell{Indian\\Pines}} & 1 & Alfalfa & 46 \\
    ~ & 2 & Corn-notill & 1428 \\
    ~ & 3 & Corn-mintill & 830 \\
    ~ & 4 & Corn & 237 \\
    ~ & 5 & Grass-pasture & 483 \\
    ~ & 6 & Grass-trees & 730 \\
    ~ & 7 & Grass-pasture-mowed & 28 \\
    ~ & 8 & Hay-windrowed & 478 \\
    ~ & 9 & Oats & 20 \\
    ~ & 10 & Soybean-notill & 972 \\
    ~ & 11 & Soybean-mintill & 2455 \\
    ~ & 12 & Soybean-clean & 593 \\
    ~ & 13 & Wheat & 205 \\
    ~ & 14 & Woods & 1265 \\
    ~ & 15 & Buildings-Grass-Trees-Drives & 386 \\
    ~ & 16 & Stone-Steel-Towers & 93 \\
    \cline{2-4}
    ~ & \multicolumn{2}{c|}{total} & 10249 \\
    \hline\hline
    \multirow{10}*{\makecell{Pavia\\University}} & 1 & Asphalt & 6631 \\
    ~ & 2 & Meadows & 18649 \\
    ~ & 3 & Gravel & 2099 \\
    ~ & 4 & Trees & 3064 \\
    ~ & 5 & Painted metal sheets & 1345 \\
    ~ & 6 & Bare Soil & 5029 \\
    ~ & 7 & Bitumen & 1330 \\
    ~ & 8 & Self-Blocking Bricks & 3682 \\
    ~ & 9 & Shadows & 947 \\
    \cline{2-4}
    ~ & \multicolumn{2}{c|}{total} & 42776 \\
    \hline\hline
    \multirow{16}*{\makecell{University\\of\\Houston}} & 1 & Healthy grass & 1251 \\
    ~ & 2 & Stressed grass & 1254 \\
    ~ & 3 & Synthetic grass & 697 \\
    ~ & 4 & Trees & 1244 \\
    ~ & 5 & Soil & 1242 \\
    ~ & 6 & Water & 325 \\
    ~ & 7 & Residential & 1268 \\
    ~ & 8 & Commercial & 1244 \\
    ~ & 9 & Road & 1252 \\
    ~ & 10 & Highway & 1227 \\
    ~ & 11 & Railway & 1235 \\
    ~ & 12 & Parking Lot 1 & 1233 \\
    ~ & 13 & Parking Lot 2 & 469 \\
    ~ & 14 & Tennis Court & 428 \\
    ~ & 15 & Running Track & 660 \\
    \cline{2-4}
    ~ & \multicolumn{2}{c|}{total} & 15029 \\
    \bottomrule[1.5pt]
    \end{tabular}
    \end{center}
\end{table}

\begin{figure}
    \centering
    \subfloat[]{
    \includegraphics[width=0.3\linewidth]{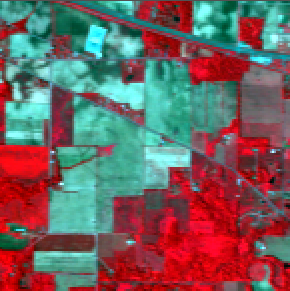}
    \label{subfig:falsecolor-indian}
    }
    \subfloat[]{
    \includegraphics[width=0.3\linewidth]{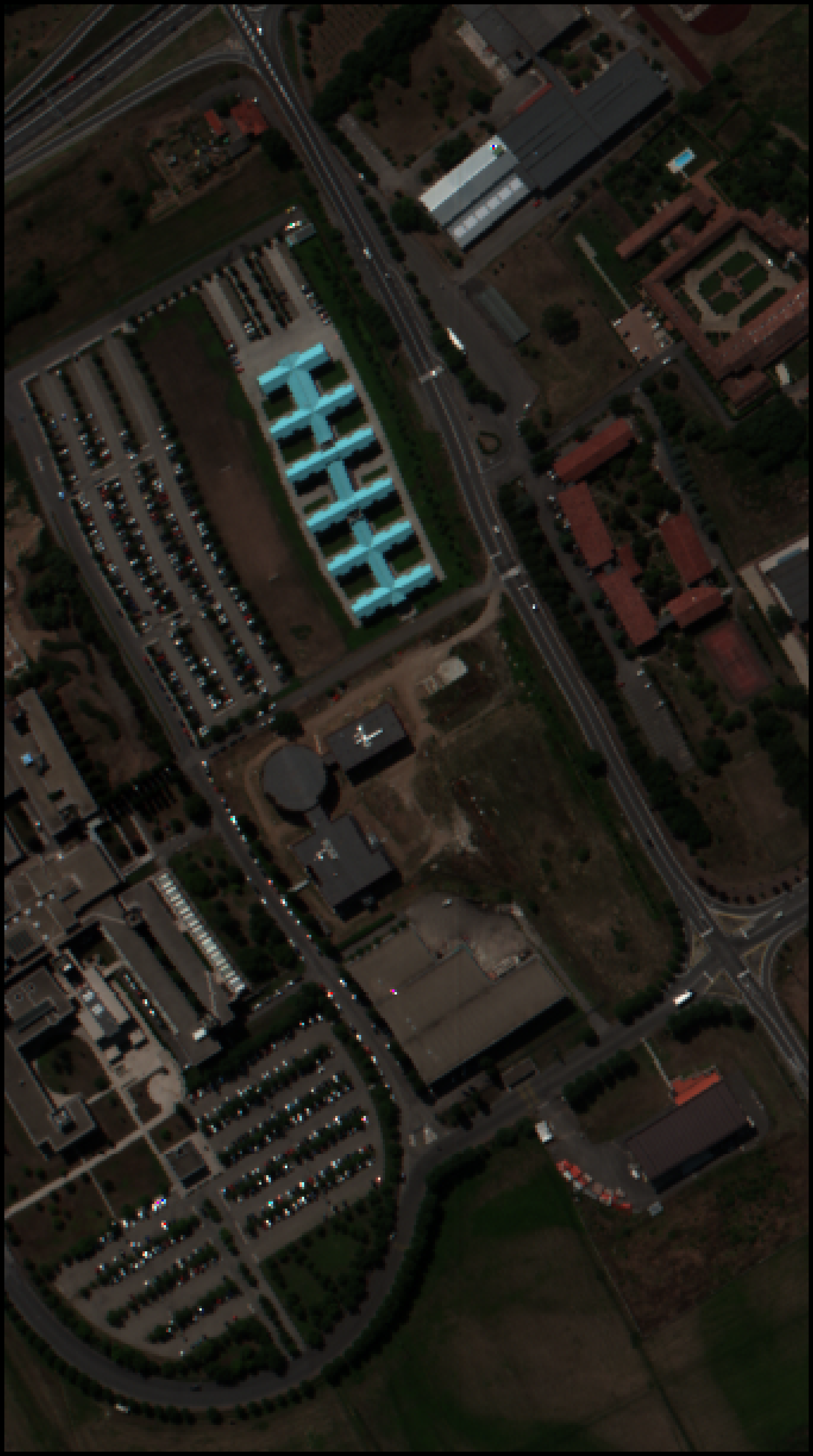}
    }
    \vfill
    \subfloat[]{
    \includegraphics[width=0.9\linewidth]{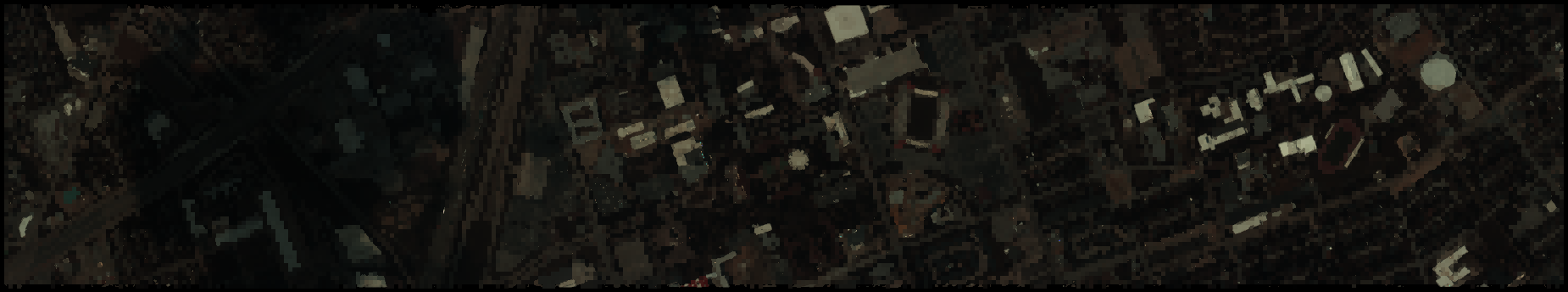}
    }
    \caption{The false-color images composited by three different spectral bands from (a) Indian Pines, (b) Pavia University, and (c) University of Houston.}
    \label{fig:falsecolor}
\end{figure}

Our crafted experiments involve a total of 6 hyperspectral datasets, namely Indian Pines, Pavia University, University of Houston, Salinas, Washington DC Mall and Kennedy Space Center.
Following the same practice of previous works, all samples are randomly divided into a training set and a testing set at a ratio of 5:95 for Indian Pines and University of Houston, and a ratio of 3:97 for Pavia University.
All training sets are further subdivided into support sets and query sets for meta-learning optimization at a unified ratio of 3:7, which guarantees that all categories are present during training.

\subsubsection{Indian Pines}

Being a publicly recognized hyperspectral band selection dataset, Indian Pines was collected in 1992 by the Airborne Visual InfraRed Imaging Spectrometer (AVIRIS) over a landscape in Indiana, USA. %
Technically, this dataset consists of a hyperspectral image with 145$\times$145 pixels.

With a wavelength range of 0.4-2.5 \si{\micro\metre}, the airborne visible infrared imager captures a total of 224 spectral reflectance bands, and discards 24 bands that cover the region of water absorption.

16 categories are involved as in Table~\ref{tab:categories}, and a false-color image composited by the bands 50, 27 and 17 is shown in Fig.~\ref{fig:falsecolor}~(a).

\subsubsection{Pavia University}

As a partial scene of Pavia, northern Italy, this dataset was acquired by the German Airborne Reflective Optical Spectral Imager Hyperspectral Data Spectral Imager (ROSIS) in 2003.
It contains 115 consecutive bands, 12 bands of which are previously removed due to noise interference.
The hyperspectral image has a spatial resolution of 610$\times$340 and contains a total of 2,207,400 pixels, including a large number of background pixels.
9 land cover categories are differenciated for classification as in Table~\ref{tab:categories}, and Fig.~\ref{fig:falsecolor}~(b) shows the extracted bands 53, 31 and 8.

\subsubsection{University of Houston}

This dataset was obtained by the ITRES CASI-1500 sensor over a campus and its neighboring urban area and served as a contest for the 2013 IEEE GRSS Data Fusion Competition.
It is a hyperspectral image of size 349$\times$1905 and contains 144 bands in the spectral range from 364 \si{\nano\metre} to 1046 \si{\nano\metre}.
Table~\ref{tab:categories} makes a list of all 15 ground object categories and their corresponding numbers of samples.
Fig.~\ref{fig:falsecolor} (c) displays a false-color image with pixel-level ground truths composed of the bands 28, 45, and 65.

\subsubsection{Salinas}

This scene was gathered by the AVIRIS sensor over Salinas Valley, California with a high spatial resolution of 512$\times$217.
Same as Indian Pines described before, only 204 spectral bands are preserved by discarding 20 bands covering water absorption.
This dataset incorporates 16 categories such as grapes, stubble and celery.

\subsubsection{Washington DC Mall}

This 191-band dataset was taken by the Hyperspectral Digital Imagery Collection Experiment (HYDICE) sensor with a wavelength range of 0.4-2.4 \si{\micro\metre}.
The hyperspectral image contains 750$\times$307 pixels, each of which is classified into one of 9 categories such as roof and grass.

\subsubsection{Kennedy Space Center}

Generally abbreviated as KSC, this dataset was collected by the AVIRIS sensor in 1996.
The spatial resolution of this dataset is 512$\times$616.
The number of spectral bands is also cut down from 224 to 176 by removing both water absorption and low SNR bands.
To achieve pixel-level classification, land cover objects are grouped into 13 categories, e.g. scrub, slash pine and salt marsh.

\begin{table*}[!t]
\caption{Classification Results of M$^3$BS and 9 Other Algorithms on a Single Dataset Indian Pines}
\label{tab:cls-single-indian}
    \begin{center}
    \begin{tabular}{c|c|c|c|c|c|c|c|c|c|c}
    \toprule[1.5pt]
    Category & \makecell{mRMR\\\cite{Peng2003FeatureSB}} & \makecell{BS-Nets\\\cite{Cai2019BSNetsAE}} & \makecell{GCSR-BS\\\cite{Cai2020EfficientGC}} & \makecell{DRLBS\\\cite{Mou2021DeepRL}} & \makecell{DGLAnet\\\cite{He2022ADG}} & \makecell{TSC\\\cite{Huang2022HeterogeneousRT}} & \makecell{SICNN\\\cite{Ghamisi2016ASC}} & \makecell{ABCNN\\\cite{RibaltaLorenzo2020HyperspectralBS}} & \makecell{BHCNN\\\cite{Feng2020ConvolutionalNN}} & M$^3$BS \\
    \midrule[1.0pt]
    OA (\%) & 71.7$\pm$0.6 & 68.7$\pm$0.9 & 68.3$\pm$2.5 & 75.2$\pm$1.2 & 81.7$\pm$0.4 & 71.4$\pm$0.7 & 87.6$\pm$1.0 & 81.5$\pm$1.7 & 96.1$\pm$0.7 & \deepen{96.5$\pm$0.4} \\
    AA (\%) & 63.1$\pm$3.4 & 59.9$\pm$1.8 & 74.6$\pm$1.1 & 67.4$\pm$1.6 & 79.9$\pm$1.0 & 67.3$\pm$3.2 & 82.7$\pm$2.6 & 74.2$\pm$2.0 & 92.2$\pm$1.3 & \deepen{95.3$\pm$1.3} \\
    Kappa$\times$100 & 67.6$\pm$0.7 & 64.3$\pm$1.0 & 71.0$\pm$1.2 & 71.7$\pm$1.4 & 79.1$\pm$0.1 & 67.3$\pm$0.7 & 85.9$\pm$1.1 & 78.8$\pm$1.4 & 95.6$\pm$1.0 & \deepen{95.9$\pm$0.3} \\
    \bottomrule[1.5pt]
    \end{tabular}
    \end{center}
\end{table*}

\begin{table*}[!ht]
\caption{Classification Results of M$^3$BS and 9 Other Algorithms on a Single Dataset Pavia University}
\label{tab:cls-single-pavia}
    \begin{center}
    \begin{tabular}{c|c|c|c|c|c|c|c|c|c|c}
    \toprule[1.5pt]
    Category & \makecell{mRMR\\\cite{Peng2003FeatureSB}} & \makecell{BS-Nets\\\cite{Cai2019BSNetsAE}} & \makecell{GCSR-BS\\\cite{Cai2020EfficientGC}} & \makecell{DRLBS\\\cite{Mou2021DeepRL}} & \makecell{DGLAnet\\\cite{He2022ADG}} & \makecell{TSC\\\cite{Huang2022HeterogeneousRT}} & \makecell{SICNN\\\cite{Ghamisi2016ASC}} & \makecell{ABCNN\\\cite{RibaltaLorenzo2020HyperspectralBS}} & \makecell{BHCNN\\\cite{Feng2020ConvolutionalNN}} & M$^3$BS \\
    \midrule[1.0pt]
    OA (\%) & 80.5$\pm$0.5 & 79.8$\pm$0.9 & 82.6$\pm$0.5 & 88.4$\pm$0.4 & 91.9$\pm$0.2 & 83.9$\pm$0.3 & 95.3$\pm$0.9 & 91.3$\pm$0.1 & 99.0$\pm$0.2 & \deepen{99.5$\pm$0.2} \\
    AA (\%) & 73.8$\pm$0.3 & 71.6$\pm$2.0 & 86.6$\pm$0.4 & 85.6$\pm$0.6 & 89.3$\pm$0.5 & 80.8$\pm$0.7 & 93.1$\pm$0.7 & 88.1$\pm$0.4 & 98.6$\pm$0.3 & \deepen{99.1$\pm$0.1} \\
    Kappa$\times$100 & 73.5$\pm$0.6 & 72.2$\pm$1.4 & 82.1$\pm$0.6 & 84.4$\pm$0.6 & 89.3$\pm$0.1 & 78.5$\pm$0.4 & 93.7$\pm$0.7 & 88.5$\pm$1.3 & 98.7$\pm$0.2 & \deepen{99.3$\pm$0.2} \\
    \bottomrule[1.5pt]
    \end{tabular}
    \end{center}
\end{table*}

\begin{table*}[!t]
\caption{Classification Results of M$^3$BS and 9 Other Algorithms on a Single Dataset University of Houston}
\label{tab:cls-single-houston}
    \begin{center}
    \begin{tabular}{c|c|c|c|c|c|c|c|c|c|c}
    \toprule[1.5pt]
    Category & \makecell{mRMR\\\cite{Peng2003FeatureSB}} & \makecell{BS-Nets\\\cite{Cai2019BSNetsAE}} & \makecell{GCSR-BS\\\cite{Cai2020EfficientGC}} & \makecell{DRLBS\\\cite{Mou2021DeepRL}} & \makecell{DGLAnet\\\cite{He2022ADG}} & \makecell{TSC\\\cite{Huang2022HeterogeneousRT}} & \makecell{SICNN\\\cite{Ghamisi2016ASC}} & \makecell{ABCNN\\\cite{RibaltaLorenzo2020HyperspectralBS}} & \makecell{BHCNN\\\cite{Feng2020ConvolutionalNN}} & M$^3$BS \\
    \midrule[1.0pt]
    OA(\%) & 82.2$\pm$0.2 & 83.3$\pm$0.5 & 88.1$\pm$0.5 & 89.7$\pm$0.6 & 91.4$\pm$0.3 & 83.8$\pm$0.3 & 93.8$\pm$0.7 & 91.6$\pm$0.4 & 96.5$\pm$0.5 & \deepen{96.7$\pm$0.4} \\
    AA(\%) & 82.0$\pm$0.2 & 82.3$\pm$0.7 & 87.1$\pm$0.5 & 88.6$\pm$0.7 & 90.3$\pm$0.6 & 83.6$\pm$0.2 & 92.7$\pm$1.7 & 91.2$\pm$0.9 & 96.4$\pm$0.6 & \deepen{96.8$\pm$0.3} \\
    Kappa$\times$100 & 80.8$\pm$0.3 & 82.0$\pm$0.6 & 87.1$\pm$0.5 & 88.9$\pm$0.7 & 90.8$\pm$0.4 & 82.9$\pm$0.3 & 93.4$\pm$0.9 & 90.9$\pm$0.8 & 96.2$\pm$0.5 & \deepen{96.5$\pm$0.4} \\
    \bottomrule[1.5pt]
    \end{tabular}
    \end{center}
\end{table*}

\subsection{Experimental Setups}
\label{subsec:exp-setting}

To more convincingly demonstrate the effectiveness of our algorithm in various aspects, up to 9 existing band selection algorithms are chosen for performance comparison, including mRMR~\cite{Peng2003FeatureSB}, BS-Nets~\cite{Cai2019BSNetsAE}, GCSR-BS~\cite{Cai2020EfficientGC}, DRLBS~\cite{Mou2021DeepRL}, DGLAnet~\cite{He2022ADG}, TSC~\cite{Huang2022HeterogeneousRT}, SICNN~\cite{Ghamisi2016ASC}, ABCNN~\cite{RibaltaLorenzo2020HyperspectralBS} and BHCNN~\cite{Feng2020ConvolutionalNN}.
Among all these methods, mRMR is a traditional filter-based method.
TSC is a recently published band selection algorithm based on subspace clustering.
The rest of 9 comparison algorithms are either based on deep learning or partially assisted with deep learning technologies.
Among them, GCSR-BS and DRLBS introduce GCN and deep reinforcement learning for a more robust representation of spectral bands in HSIs.
BS-Nets, DGLAnet, SICNN, ABCNN and BHCNN are 5 modern CNN-based methods with strong spatial representation capacity.

The methods, mRMR, GCSR-BS and TSC, require additional classifiers.
Here, a SVM classifier is attached by grid searching $c$ and $\gamma$ in the range of \{1, 10, 100, 1000, 10000\} and \{0.01, 0.1, 1, 10, 100\}, respectively.
For BS-Nets with a trainable CNN, the learning rate is fixed to 0.001, the number of training epochs is set to 500, and the regularization coefficient is set to 0.02.
For deep learning-based methods, BS-Nets, GCSR-BS, DRLBS, DGLAnet, SICNN, ABCNN and BHCNN, the hyperparameters such as batch size, learning rate and training epochs are determined either with a trial-and-error procedure or by following the settings of the original paper.

For our M$^3$BS, both the learning rate $\lr$ for optimizing classification and the meta-learning rate $\mlr$ for optimizing band selection are set to 0.001, and further adopt an exponential scheduler to decay themselves every epoch.
Both the generalizable GCN and the CNN-based image classifiers are optimized with a gradient-based stochastic algorithm Adam~\cite{Kingma2014AdamAM}.
The batch size $\batchsize$ is set to 128, and the number of epochs $\nepoch$ is set to 400.
For any batch of any dataset, the spatial size of the hyperspectral image patch is fixed to 33$\times$33.
By controlling the truncation threshold $\adjthres$ for the adjacency matrix $\adjmat$, the number of edges is limited to less than 1000 to prevent over-densification.
The number of dataset-agnostic bases for parameterizing the graph convolutional kernels is set to 3.
After extracting importance scores for each band, the number of selected bands $\nsband$ is fixed to 20 for a fair comparison among different datasets.
From a diversity perspective, BS-Nets, SICNN and TWCNN, which represents filter-based, wrapper-based and embedding-based methods respectively, are chosen to be our teachers.
For the sake of reproducibility, detailed experimental procedures and well-commented implementations in PyTorch~\cite{Paszke2019PyTorchAI}.
All the experiments are conducted on a computer with a AMD Ryzen 5950X CPU and a Nvidia GeForce RTX 3090 GPU.

\subsection{Classification Results on a Single Dataset}
\label{subsec:exp-cls-single}

In this subsection, to preliminarily confirm the effectiveness of our network structure, we verify the classification performance of all the comparison algorithms.
This scenario has nothing to do with zero-shot learning.
Three well-known classification metrics, named the overall accuracy (OA), the average accuracy (AA), and the Kappa coefficient (Kappa), are used for quantitative comparison of the classification performance among different algorithms.

\subsubsection{Indian Pines}

The dataset is randomly divided into 5\% training set and 95\% testing set, and a subset consisting of 20 bands is selected from the original 200 spectral bands.
Table~\ref{tab:cls-single-indian} records the classification results of each algorithm, including the OA, AA and Kappa. The classification accuracy of each class is available  Table S1 in the Supplementary Material.
The corresponding standard deviations of each accuracy is given by running each algorithm 30 times independently.
Among all these results from 10 competitive algorithms, the best ones are bolded and highlighted in gray.
Among all these algorithms, with an OA below 70\%, BS-Nets and GCSR-BS have the worst classification performance.
The former formalizes band selection as an spectral reconstruction task, and the latter does not utilize any label information for classification.
As band selection methods with auxiliary classifiers, mRMR, DRLBS and TSC have slightly better results due to their reasonable optimization strategies.
However, for these algorithms, the classification is acheived by an independent SVM and optimized separately from the band selection task, which hinders mutual promotion between different tasks.
By integrating an attention module for richer spatial-spectral and global-local features, DGLAnet receives nearly 10\% performance improvement.
Thanks to the spatial representation and nonlinear fitting ability of CNNs, the CNN-based algorithms, including SICNN, ABCNN, BHCNN and M$^3$BS, are significantly better than the previous ones.
Among them, the accuracy of BHCNN is about 10\% greater than that of SICNN due to its bandwise-independent convolution layers, which are more suitable for band selection.
Finally, our M$^3$BS acheives the best accuracy on most categories and on all the three classification metrics.

\subsubsection{Pavia University}

The dataset is splitted into 3\% training samples and 97\% testing samples.
For all competitive algorithms, 20 bands are selected from the original 103 bands.
Table~\ref{tab:cls-single-pavia} demonstrates the quantitative comparison of classification ability among M$^3$BS and 9 other algorithms. The classification accuracy of each class is available  Table S2 in the Supplementary Material.
Similar to the situation in Indian Pines, end-to-end deep learning algorithms including SICNN, ABCNN, BHCNN and M$^3$BS, have an accuracy improvement of 5\%-15\% compared to traditional algorithms such as mRMR and GCSR-BS.
Meanwhile, among all deep learning algorithms, M$^3$BS achieves the best results on most of the 9 categories.
Compared with the representative CNN-based algorithm SICNN, M$^3$BS gains an improvement over the overall accuracy by 4.1\%, the average accuracy by 5.7\%, and the Kappa coefficient by 5.2\%.
In addition, it is difficult for most algorithms to correctly classify the three categories Gravel, Trees and Self-Blocking Bricks, while M$^3$BS improves these categories by at most 69.7\%, 21.5\% and 22.4\%, and at least 1.1\%, 0.5\% and 0.7\% respectively.

\begin{table*}[!ht]
\caption{Zero-Shot Classification Results of M$^3$BS and 9 Other Algorithms on the Unseen Dataset Indian Pines}
\label{tab:cls-zeroshot-indian}
    \begin{center}
    \begin{tabular}{c|c|c|c|c|c|c|c|c|c|c}
    \toprule[1.5pt]
    ~ & \makecell{mRMR\\\cite{Peng2003FeatureSB}} & \makecell{BS-Nets\\\cite{Cai2019BSNetsAE}} & \makecell{GCSR-BS\\\cite{Cai2020EfficientGC}} & \makecell{DRLBS\\\cite{Mou2021DeepRL}} & \makecell{DGLAnet\\\cite{He2022ADG}} & \makecell{TSC\\\cite{Huang2022HeterogeneousRT}} & \makecell{SICNN\\\cite{Ghamisi2016ASC}} & \makecell{ABCNN\\\cite{RibaltaLorenzo2020HyperspectralBS}} & \makecell{BHCNN\\\cite{Feng2020ConvolutionalNN}} & M$^3$BS \\
    \midrule[1.0pt]
    OA (\%) & 70.8$\pm$0.8 & 74.0$\pm$0.3 & 70.5$\pm$1.4 & 72.6$\pm$1.2 & 78.6$\pm$0.8 & 69.3$\pm$2.6 & 85.0$\pm$3.2 & 77.6$\pm$2.1 & 90.5$\pm$1.3 & \deepen{96.2$\pm$0.5} \\
    AA (\%) & 60.9$\pm$1.3 & 64.3$\pm$1.6 & 65.5$\pm$2.6 & 60.0$\pm$1.5 & 76.9$\pm$1.7 & 63.5$\pm$2.4 & 78.6$\pm$3.4 & 70.6$\pm$3.4 & 78.6$\pm$2.5 & \deepen{94.3$\pm$0.9} \\
    Kappa$\times$100 & 66.6$\pm$0.9 & 70.0$\pm$0.3 & 66.2$\pm$1.6 & 68.5$\pm$1.4 & 76.4$\pm$1.3 & 63.6$\pm$1.4 & 82.8$\pm$3.7 & 75.4$\pm$2.2 & 89.1$\pm$1.5 & \deepen{95.7$\pm$0.5} \\
    \bottomrule[1.5pt]
    \end{tabular}
    \end{center}
\end{table*}

\begin{table*}[!ht]
\caption{Zero-Shot Classification Results of M$^3$BS and 9 Other Algorithms on the Unseen Dataset Pavia University}
\label{tab:cls-zeroshot-pavia}
    \begin{center}
    \begin{tabular}{c|c|c|c|c|c|c|c|c|c|c}
    \toprule[1.5pt]
    ~ & \makecell{mRMR\\\cite{Peng2003FeatureSB}} & \makecell{BS-Nets\\\cite{Cai2019BSNetsAE}} & \makecell{GCSR-BS\\\cite{Cai2020EfficientGC}} & \makecell{DRLBS\\\cite{Mou2021DeepRL}} & \makecell{DGLAnet\\\cite{He2022ADG}} & \makecell{TSC\\\cite{Huang2022HeterogeneousRT}} & \makecell{SICNN\\\cite{Ghamisi2016ASC}} & \makecell{ABCNN\\\cite{RibaltaLorenzo2020HyperspectralBS}} & \makecell{BHCNN\\\cite{Feng2020ConvolutionalNN}} & M$^3$BS \\
    \midrule[1.0pt]
    OA (\%) & 73.4$\pm$0.4 & 79.5$\pm$0.4 & 82.6$\pm$0.3 & 86.2$\pm$0.7 & 87.3$\pm$0.5 & 79.7$\pm$1.2 & 94.5$\pm$0.6 & 81.5$\pm$1.1 & 94.2$\pm$1.6 & \deepen{99.5$\pm$0.2} \\
    AA (\%) & 63.2$\pm$1.0 & 66.7$\pm$1.1 & 76.6$\pm$0.7 & 79.5$\pm$2.9 & 86.1$\pm$0.7 & 77.6$\pm$1.1 & 90.4$\pm$1.6 & 80.9$\pm$2.3 & 89.8$\pm$3.6 & \deepen{98.5$\pm$0.3} \\
    Kappa$\times$100 & 63.3$\pm$0.5 & 71.2$\pm$0.6 & 76.2$\pm$0.4 & 81.2$\pm$1.0 & 86.2$\pm$0.7 & 77.3$\pm$0.9 & 92.7$\pm$0.9 & 81.1$\pm$1.7 & 92.3$\pm$2.1 & \deepen{99.2$\pm$0.2} \\
    \bottomrule[1.5pt]
    \end{tabular}
    \end{center}
\end{table*}

\begin{table*}[!ht]
\caption{Zero-Shot Classification Results of M$^3$BS and 9 Other Algorithms on the Unseen Dataset University of Houston}
\label{tab:cls-zeroshot-houston}
    \begin{center}
    \begin{tabular}{c|c|c|c|c|c|c|c|c|c|c}
    \toprule[1.5pt]
    ~ & \makecell{mRMR\\\cite{Peng2003FeatureSB}} & \makecell{BS-Nets\\\cite{Cai2019BSNetsAE}} & \makecell{GCSR-BS\\\cite{Cai2020EfficientGC}} & \makecell{DRLBS\\\cite{Mou2021DeepRL}} & \makecell{DGLAnet\\\cite{He2022ADG}} & \makecell{TSC\\\cite{Huang2022HeterogeneousRT}} & \makecell{SICNN\\\cite{Ghamisi2016ASC}} & \makecell{ABCNN\\\cite{RibaltaLorenzo2020HyperspectralBS}} & \makecell{BHCNN\\\cite{Feng2020ConvolutionalNN}} & M$^3$BS \\
    \midrule[1.0pt]
    OA (\%) & 80.6$\pm$1.0 & 79.7$\pm$0.5 & 88.5$\pm$1.2 & 79.9$\pm$0.8 & 88.6$\pm$0.2 & 81.4$\pm$0.6 & 88.9$\pm$1.7 & 83.7$\pm$1.6 & 91.4$\pm$2.5 & \deepen{96.5$\pm$0.5} \\
    AA (\%) & 79.5$\pm$1.1 & 79.2$\pm$0.4 & 87.4$\pm$1.2 & 78.9$\pm$0.9 & 87.0$\pm$0.5 & 81.1$\pm$0.2 & 87.3$\pm$1.9 & 83.0$\pm$1.3 & 90.8$\pm$2.3 & \deepen{95.9$\pm$0.3} \\
    Kappa$\times$100 & 79.0$\pm$1.1 & 78.1$\pm$0.5 & 87.6$\pm$1.2 & 78.2$\pm$0.8 & 87.9$\pm$0.6 & 80.7$\pm$0.3 & 88.0$\pm$1.8 & 82.4$\pm$1.8 & 90.6$\pm$2.7 & \deepen{96.3$\pm$0.4} \\
    \bottomrule[1.5pt]
    \end{tabular}
    \end{center}
\end{table*}

\subsubsection{University of Houston}

Same as Indian Pines, the dataset is divided into a training set and a testing set at a ratio of 5:95.
For all 10 algorithms for comparison, 20 bands among 176 bands are selected for hyperspectral band selection.
Table~\ref{tab:cls-single-houston} lists the classification accuracy of each algorithm on the entire dataset. The classification accuracy of each class is available  Table S3 in the Supplementary Material.
Our M$^3$BS performs best on most of the 15 categories, and maintains a classification accuracy above 94\% on almost all the categories.
Compared with ABCNN, M$^3$BS exhibits at least 5.1\%, 5.6\% and 5.6\% in terms of OA, AA and Kappa, respectively.
For some hard-to-classify categories, such as Commercial and Road, M$^3$BS maintains an accuracy of more than 93\%, while the accuracy of almost all other competitive algorithms is below 90\%.

\begin{table*}[!t]
\caption{Classification Results of M$^3$BS with different numbers of teachers on different datasets}
\label{tab:ablation-teachers}
    \begin{center}
    \begin{tabular}{c|l|c|c|c}
    \toprule[1.5pt]
    Dataset & Condition & OA (\%) & AA (\%) & Kappa$\times$100 \\
    \midrule[1.0pt]
    \multirow{4}*{\makecell{Indian Pines}} & M$^3$BS w/o teachers & 93.9 & 90.3 & 93.1 \\
    ~ & M$^3$BS w/ 1 teacher (BS-Nets) & 95.6 & 91.4 & 95.0 \\
    ~ & M$^3$BS w/ 2 teachers (BS-Nets and SICNN) & 95.8 & 92.3 & 95.2 \\
    ~ & M$^3$BS w/ 3 teachers (BS-Nets, SICNN and TWCNN) & \deepen{96.2} & \deepen{94.3} & \deepen{95.7} \\
    \hline\hline
    \multirow{4}*{\makecell{Pavia University}} & M$^3$BS w/o teachers & 97.2 & 95.7 & 96.4 \\
    ~ & M$^3$BS w/ 1 teacher (BS-Nets) & 98.2 & 96.7 & 97.7 \\
    ~ & M$^3$BS w/ 2 teachers (BS-Nets and SICNN) & 98.4 & 97.1 & 97.9 \\
    ~ & M$^3$BS w/ 3 teachers (BS-Nets, SICNN and TWCNN) & \deepen{98.5} & \deepen{97.5} & \deepen{98.1} \\
    \hline\hline
    \multirow{4}*{\makecell{University of Houston}} & M$^3$BS w/o teachers & 93.4 & 92.9 & 93.0 \\
    ~ & M$^3$BS w/ 1 teacher (BS-Nets) & 95.0 & 94.5 & 94.7 \\
    ~ & M$^3$BS w/ 2 teachers (BS-Nets and SICNN) & 95.9 & 95.3 & 95.6 \\
    ~ & M$^3$BS w/ 3 teachers (BS-Nets, SICNN and TWCNN) & \deepen{96.5} & \deepen{95.9} & \deepen{96.3} \\
    \bottomrule[1.5pt]
    \end{tabular}
    \end{center}
\end{table*}

\begin{table}[!ht]
\caption{Classification Results of M$^3$BS with different multi-teacher fusion strategies on different datasets}
\label{tab:ablation-fusion}
    \begin{center}
    \begin{tabular}{c|l|c|c|c}
    \toprule[1.5pt]
    Dataset & Condition & OA (\%) & AA (\%) & Kappa$\times$100 \\
    \midrule[1.0pt]
    \multirow{3}*{\makecell{Indian\\Pines}} & union & 95.4 & 92.4 & 94.8 \\
    ~ & normalized sum & 96.1 & 91.7 & 95.5 \\
    ~ & diversity ensemble & \deepen{96.2} & \deepen{94.3} & \deepen{95.7} \\
    \hline\hline
    \multirow{3}*{\makecell{Pavia\\University}} & union & 98.0 & 96.7 & 97.4 \\
    ~ & normalized sum & 98.2 & 96.8 & 97.7 \\
    ~ & diversity ensemble & \deepen{98.5} & \deepen{97.5} & \deepen{98.1} \\
    \hline\hline
    \multirow{3}*{\makecell{University\\of\\Houston}} & union & 96.4 & \deepen{96.1} & 96.2 \\
    ~ & normalized sum & 92.7 & 92.6 & 92.2 \\
    ~ & diversity ensemble & \deepen{96.5} & 95.9 & \deepen{96.3} \\
    \bottomrule[1.5pt]
    \end{tabular}
    \end{center}
\end{table}

\begin{table*}[!t]
\caption{Classification Results of M$^3$BS with Different Multi-Loss Weighting Schemes on Different Datasets}
\label{tab:ablation-weighting}
    \begin{center}
    \begin{tabular}{c|c|c|c|c|c|c}
    \toprule[1.5pt]
    \multirow{2}*{Scenario} & \multirow{2}*{Dataset} & \multicolumn{2}{c}{Weights} & \multirow{2}*{OA (\%)} & \multirow{2}*{AA (\%)} & \multirow{2}*{Kappa$\times$100} \\
    ~ & ~ & $\weight^\text{bs}$ & $\weight^\text{cls}$ & ~ & ~ & ~ \\
    \midrule[1.0pt]
    \multirow{3}*{only $\loss^\text{cls}$} & Indian Pines & 0 & 1 & 93.9 & 90.3 & 93.1 \\
    ~ & Pavia University & 0 & 1 & 97.2 & 95.7 & 96.4 \\
    ~ & University of Houston & 0 & 1 & 93.4 & 92.9 & 93.0 \\
    \midrule[0.5pt]
    \multirow{21}*{\makecell{hand-crafted\\weighting}} & Indian Pines & 0.01 & 0.99 & 95.5 & 92.2 & 94.9 \\
    ~ & Pavia University & 0.01 & 0.99 & 98.1 & 96.8 & 97.6 \\
    ~ & University of Houston & 0.01 & 0.99 & 95.0 & 94.3 & 94.7 \\
    \cmidrule[0.5pt](){2-7}
    ~ & Indian Pines & 0.05 & 0.95 & 95.7 & 92.1 & 95.2 \\
    ~ & Pavia University & 0.05 & 0.95 & 98.0 & 96.7 & 97.5 \\
    ~ & University of Houston & 0.05 & 0.95 & 94.9 & 94.6 & 94.6 \\
    \cmidrule[0.5pt](){2-7}
    ~ & Indian Pines & 0.1 & 0.9 & 95.7 & 93.0 & 95.2 \\
    ~ & Pavia University & 0.1 & 0.9 & 98.0 & 96.6 & 97.5 \\
    ~ & University of Houston & 0.1 & 0.9 & 95.0 & 94.3 & 94.6 \\
    \cmidrule[0.5pt](){2-7}
    ~ & Indian Pines & 0.3 & 0.7 & 96.0 & 93.1 & 95.5 \\
    ~ & Pavia University & 0.3 & 0.7 & 98.2 & 97.1 & 97.8 \\
    ~ & University of Houston & 0.3 & 0.7 & 95.1 & 94.5 & 94.8 \\
    \cmidrule[0.5pt](){2-7}
    ~ & Indian Pines & 0.7 & 0.3 & 96.2 & 93.3 & 95.7 \\
    ~ & Pavia University & 0.7 & 0.3 & 98.2 & 96.9 & 97.7 \\
    ~ & University of Houston & 0.7 & 0.3 & 95.0 & 94.5 & 94.7 \\
    \cmidrule[0.5pt](){2-7}
    ~ & Indian Pines & 0.9 & 0.1 & 95.9 & 92.5 & 95.3 \\
    ~ & Pavia University & 0.9 & 0.1 & 98.2 & 96.8 & 97.7 \\
    ~ & University of Houston & 0.9 & 0.1 & 95.0 & 94.5 & 94.7 \\
    \midrule[0.5pt]
    \multirow{3}*{\makecell{averaged\\weighting}} & Indian Pines & 0.5 & 0.5 & 95.6 & 92.3 & 95.0 \\
    ~ & Pavia University & 0.5 & 0.5 & 98.3 & 97.0 & 97.9 \\
    ~ & University of Houston & 0.5 & 0.5 & 95.0 & 94.4 & 94.7 \\
    \midrule[0.5pt]
    \multirow{3}*{\makecell{uncertainty\\weighting}} & Indian Pines & \multicolumn{2}{c|}{\multirow{3}*{auto}} & \deepen{96.2} & \deepen{94.3} & \deepen{95.7} \\
    ~ & Pavia University & \multicolumn{2}{c|}{} & \deepen{98.5} & \deepen{97.5} & \deepen{98.1} \\
    ~ & University of Houston & \multicolumn{2}{c|}{} & \deepen{96.5} & \deepen{95.9} & \deepen{96.3} \\
    \bottomrule[1.5pt]
    \end{tabular}
    \end{center}
\end{table*}

\subsection{Zero-Shot Classification Results on the Unseen Dataset}
\label{subsec:exp-cls-zeroshot}

What we initially want to implement is zero-shot band selection, that is, selecting spectral bands for a new dataset without any of its samples during training.
To verify the performance of our M$^3$BS in this scenario, we introduce three additional hyperspectral datasets, namely Salinas, Washington DC Mall and KSC, to be meta-training tasks, and regard the above-mentioned Indian Pines, Pavia University and University of Houston datasets as meta-testing tasks.
Similar to the practice of previous datasets, these meta-training datasets are splited into training sets and testing sets at a ratio of 10:90.
In other words, what we want to achieve is to train our algorithm only with three datasets and directly perform band selection on the other three datasets.
It should be noted that, this scenario differs from unsupervised learning, in which only labels rather than samples are absent.

None of the 9 comparison algorithms are optimized for zero-shot learning, and some of them are not compatible with multiple training datasets.
To make training feasible, we apply some intuitive modifications to these algorithms.
For multiple datasets to be trained on, we firstly find the dataset containing the most spectral bands, and then fill in the missing bands with 0 for other datasets with less bands.
As a result, all training datasets are aligned in terms of the number of spectral bands.
For mRMR, after calculating the score of the selected band combination for each training dataset, the average of these scores are taken to be the final score for further updates.
For BS-Nets, we set up as many reconstruction networks as there are datasets, and take the average of the output of these networks as the final result.
For CNN-based algorithms SICNN and ABCNN, their CNN-based classifiers are retrained for different band subsets during the training phase.
For BHCNN, multiple independent classifiers instead of one classifier are attached for parallel optimization.

It should be pointed out that, since our M$^3$BS is originally designed for zero-shot learning, it is naturally compatible with multiple training datasets and does not require an additional dataset alignment operation.
Taking the first graph convolutional layer as an example, the size $\height\width \times \dhid$ of the weight matrix $\wgtmat^\text{GCN(1)}$ is independent of the number of bands $\nband$ for any HSI dataset.
As long as the patch size $\height \times \width$ is fixed, the same learnable parameters are appliable to datasets with any number of bands.
The relationship between the network structure and the amount of bands is decoupled in M$^3$BS.

Table~\ref{tab:cls-zeroshot-indian}-\ref{tab:cls-zeroshot-houston} shows the average of the classification metrics OA, AA and Kappa and their corresponding standard deviations on three datasets by 30 independent runs of each algorithm.
Among these accuracies, the best ones are bolded and highlighted as in Table~\ref{tab:cls-single-indian}-\ref{tab:cls-single-houston}.
It can be seen at a glance from these tables that, our M$^3$BS significantly outperforms all the other 9 algorithms, and maintains a consistently high level over all three datasets that are absent throughout the whole training phase.
Quantitatively, for any of these meta-testing datasets, M$^3$BS can improve at least 4.3\% OA, 5.1\% AA, and 5.7\% Kappa compared to the best performing one among other algorithms.
This is predominantly because both the network structure and the optimization procedure in M$^3$BS are designed for adapting to zero-shot band selection.
Specifically, the generalizable GCN with the additional supervision from multiple teachers can achieve meta-knowledge extraction that is compatible with various datasets.
The multi-objective meta-learning optimization procedure makes M$^3$BS appliable to different objectives and different datasets.
As a result, the extracted meta-knowledge can be directly transferred to unseen datasets without any retraining or fine-tuning.
In contrast, the knowledge learned by other comparison algorithms is not transferable, which means that these algorithms perform well on seen datasets but poorly on unseen datasets.

For a more intuitive comparison, the ground truth (G.T.) and the visual classification results on three unseen hyperspectral datasets of these algorithms are shown in Fig.S1-Fig.S3 in the Supplementary Material.
It should be noted that, the full-image classification predictions of the dataset University of Houston are given due to the scattered distribution of labeled pixel samples.
As shown in (b-j), a large number of pixel samples are misclassified into other categories.
For mRMR, BS-Nets, GCSR-BS, DRLBS, TSC and ABCNN, the network input is constructed at the pixel level, which does not consider the relationship among adjacent pixels, and lacks local spatial information.
As a result, the output of these methods contains lots of noise points and discontinuous areas, espacially for categories like Grass-trees, Soybean-notill, Soybean-mintill in Indian Pines and Bare Soil in Pavia University.
For CNN-based algorithms SICNN and BHCNN, the spatial network takes the spatial window around the pixel as input, which significantly improves local spatiality and category consistency in the same area.
However, misclassified pixels still appear at some boundaries among different categories.
In M$^3$BS, pixel-level samples are used in the process of constructing spatial-spectral graphs, and spatial windows are regarded as input for the classifier.
Compared with the above methods, M$^3$BS not only utilizes a more efficient graph construction strategy, but also takes local spatial information into consideration.
As a result, M$^3$BS obtains the best classification visualization among all comparison algorithms, which preserves detailed boundaries, has the least noise points and is closest to the ground truth.

\subsection{Ablation Experiments on the Unseen Dataset}
\label{subsec:exp-ablation}

\subsubsection{Multiple Teachers}

This ablation aims to verify whether multiple teachers are beneficial for band selection.
To quantitatively analyze the contribution of each teacher, Table~\ref{tab:ablation-teachers} records the classification results of M$^3$BS with different numbers of teachers.

As shown in Table~\ref{tab:ablation-teachers}, as the number of band selection teachers increases, the classification performance of M$^3$BS keeps improving.
The number of teachers being zero corresponds to the absence of both the band selection loss $\loss^\text{bs}$ and the multi-objective loss $\loss$.
At this time, the classification accuracy is about 3\% lower than the original M$^3$BS, which proves that the classification task alone can't provide sufficient meta-knowledge for zero-shot band selection.
As different teachers join, the classification ability becomes stronger and stronger since different types of teachers can provide guidance from different perspectives.
Additionally, the order in which teachers join shows that, the supervision from deep learning algorithms like SICNN and TWCNN, is of higher quality than that from unsupervised algorithms like BS-Nets.

\subsubsection{Multi-Teacher Fusion}

After confirming that multiple teachers are indeed beneficial for zero-shot band selection, this ablation try to figure out the best strategy to fuse these teachers into one integrated teacher.

Our original M$^3$BS utilizes a strategy called diversity ensemble, which chooses the spectral bands according to their popularity.
Bands with more received votes will be selected first, and bands with the same number of received votes will be randomly selected.
For comparison, we compare M$^3$BS with two intuitive fusion strategies, named union and normalized sum.
The former treats the band combinations selected by teachers as sets, and takes the union of these sets as the result.
The latter multiplies each teacher's output score by one-third, and add these scores together as the ground truth.

Table~\ref{tab:ablation-fusion} records the classification results of M$^3$BS with different multi-teacher fusion strategies.

As shown in Table~\ref{tab:ablation-fusion}, our novel fusion strategy outperforms the other two strategies with an accuracy increase of about 1\%.
For the strategy union, we suspect that this is due to the so-called over-selection problem.
In other words, when the bands selected by different teachers do not have much overlap, too many bands will be used as fusion results, and some bands with low discriminability may be added, which in turn downgrades the classification performance.
For the strategy normalized sum, we think that it may lack the introduction of random noise, thus leaving the supervision for each epoch unchanged and reducing the generalization ability.
On the contrary, the diversity ensemble strategy we proposed not only limits the number of selected bands, but also introduces a certain degree of randomness, thereby ensuring the effective fusion of multiple teachers.

\subsubsection{Multi-Loss Weighting}

This ablation is used to prove the necessity of balancing multiple losses, and to make comparison among hand-crafted and automatically learned loss weights.
When encountering multiple loss functions in a deep learning algorithm, the intuitive idea is to manually set up different weights for different losses, and keep these weights unchanged throughout the training process.
Here, we design 6 hand-crafted weight combinations in sequence.
Also as usual, the sum of all the loss weights in a single combination is fixed to 1.
Instead of manual intervention on these weights, our M$^3$BS utilizes a homoscedastic uncertainty-based multi-objective loss to automatically learn them.
For a more refined and convincing comparison, Table~\ref{tab:ablation-weighting} elaborately lists the classification results of M$^3$BS with different multi-loss weighting schemes.

As shown in Table~\ref{tab:ablation-weighting}, the uncertainty-based weighting scheme introduced in M$^3$BS dominates all other weighting schemes.
First, we do not show the classification performance when only optimizing the band selection loss $\loss^\text{bs}$.
When the classification loss $\loss^\text{cls}$ is absent, the CNN-based classifier of each dataset is not trained any more, leading to random network parameters for inferring, which makes no sense.
Next, the case of only optimizing the classification loss $\loss^\text{cls}$ is actually the same as the case without multiple teachers in Table~\ref{tab:ablation-teachers}, and the accuracy is about 3\% lower than M$^3$BS.
Here, the same assertion can be applied: a single classification task is not enough for sufficient meta-knowledge for zero-shot band selection.
Among these hand-crafted weight combinations, we can observe that for the dataset Indian Pines, Pavia University and University of Houston, the optimal weight ratios are approximately 0.7:0.3, 0.5:0.5 and 0.3:0.7, respectively.
Finally, the weight combination learned by our uncertainty weighting scheme is better than any hand-crafted weight.
It's reasonable to speculate that, since these weights are updated together with the network parameters, they can be adapted to different training stages.
For example, in the early stage, band selection is entrusted to dominate the network training, leading to a preliminary understanding of spectral bands; in the later stage, the classification takes the lead to further strengthening the feature representation ability.
This is only available for multi-objective learning, not for static weights.

\section{Conclusion}
\label{sec:concl}

In this paper, a novel multi-teacher multi-objective meta-learning framework M$^3$BS is proposed for zero-shot hyperspectral band selection.
For compatible meta-knowledge extraction for various datasets, M$^3$BS constructs a generalizable GCN with spatial-spectral graphs.
The decomposition of dataset-agnostic bases and dataset-specific coefficients makes it possible to explicitly characterize unseen samples with the unified meta-knowledge.
For a better generalization ability and a faster convergence speed, M$^3$BS ensemble multiple teachers to provide high-quality experiences.
For zero-shot optimization over multiple datasets and multiple objectives simultaneously, all these modules are integrated into a multi-objective meta-learning procedure.
Experimental results over three commonly used hyperspectral datasets prove that M$^3$BS can produce better band combinations than many other state-of-the-art baselines for zero-shot learning.
Moreover, it provides a new design paradigm for subsequent researchers, and reminds them to pay more attention to the performance when transferring to unseen datasets.

\ifCLASSOPTIONcaptionsoff
  \newpage
\fi

\bibliographystyle{IEEEtran}
\bibliography{IEEEabrv,main}

\end{document}